\documentclass[10pt,twocolumn,letterpaper]{article}

\usepackage{cvpr}              

\usepackage{enumitem}
\usepackage{algorithm}
\usepackage{algpseudocode}
\usepackage{xcolor}
\usepackage{graphicx}
\usepackage{booktabs}
\usepackage{pifont}

\newcommand{\cmark}{\ding{51}}
\newcommand{\xmark}{\ding{55}}

\floatname{algorithm}{Algorithm}

\definecolor{cvprblue}{rgb}{0.21,0.49,0.74}
\definecolor{mypink}{RGB}{220,0,120}

\usepackage[
    pagebackref,
    breaklinks,
    colorlinks,
    linkcolor=cvprblue,
    citecolor=cvprblue,
    urlcolor=mypink
]{hyperref}

\usepackage{multirow}

\usepackage{graphicx}
\usepackage{subcaption}

\def\paperID{} 
\def\confName{CVPR}
\def\confYear{2026}

\title{It's Time to Get It Right: Improving Analog Clock Reading and Clock-Hand Spatial Reasoning in Vision-Language Models}

\author{
Jaeha Choi$^{1}$\thanks{Equal contribution.} \and
Jin Won Lee$^{2}$\footnotemark[1] \and
Siwoo You$^{1}$ \and
Jangho Lee$^{1}$ \\
$^{1}$Incheon National University, Incheon, Republic of Korea \\
$^{2}$McGill University, Montreal, Canada \\
{\tt\small
\{chlgocks2000, syousiwoos, ubuntu\}@inu.ac.kr \quad
jinwon.lee@mail.mcgill.ca
} \\
\small \url{https://it-s-time-to-get-it-right.github.io/}
}

\begin{document}
\maketitle


\begin{abstract}
Advances in vision-language models (VLMs) have achieved remarkable success on complex multimodal reasoning tasks, leading to the assumption that they should also excel at reading analog clocks. However, contrary to this expectation, our study reveals that reading analog clocks in real-world environments remains a significant challenge for state-of-the-art VLMs.
Existing analog clock datasets are largely synthetic or planar with limited stylistic diversity and minimal background context, failing to capture the visual variability of real-world scenes. As a result, VLMs trained on such data exhibit weak spatiotemporal reasoning, frequently confusing the hour and minute hands and struggling under common visual conditions such as occlusion, lighting variation, and cluttered backgrounds. 
To address this issue, we introduce \textbf{TickTockVQA}, a human-annotated dataset containing analog clocks in diverse real-world scenarios.
TickTockVQA provides explicit hour and minute annotations, and includes an AM/PM tag when it is inferable from the visual context.
Furthermore, we propose \textbf{Swap-DPO}, a direct preference optimization-based fine-tuning framework to align model reasoning toward accurate time interpretation. Experimental results demonstrate that our approach substantially enhances clock reading accuracy and robustness under real-world conditions, establishing a foundation for future research on spatiotemporal reasoning and visual understanding in VLMs.
\end{abstract}
\section{Introduction} \label{sec:intro}

Reading time from an analog clock is an everyday skill for humans, yet it remains a challenging problem for modern vision-language models (VLMs).
As VLMs~\cite{molmo2024, qwen2.5vl, dubey2024llama, gemma3, internvl3} increasingly serve as the foundation for general-purpose multimodal and embodied AI systems~\cite{arora2024g2tr, zhou2025roborefer, kong2025autospatial}, their failure to robustly read analog clocks reveals a concrete limitation in their spatiotemporal reasoning abilities.
Although the task of reading an analog clock may appear narrow, the underlying capability is relevant to a wide range of real-world scenarios where temporal information is conveyed visually rather than through explicit text.
The task requires jointly locating the clock, identifying its hands, interpreting their geometric configuration, and mapping continuous angular relationships to discrete time values~\cite{saxena2025lost}.
Similar work on reading analog scales~\cite{howells2021real, reitsma2024under} has shown that handling such conditions requires multi-level spatial reasoning.
This makes analog clock reading a concise but rich testbed for studying fine-grained spatiotemporal reasoning in VLMs.
%
\begin{figure}[t]
    \centering
    \includegraphics[width=1.0\linewidth]{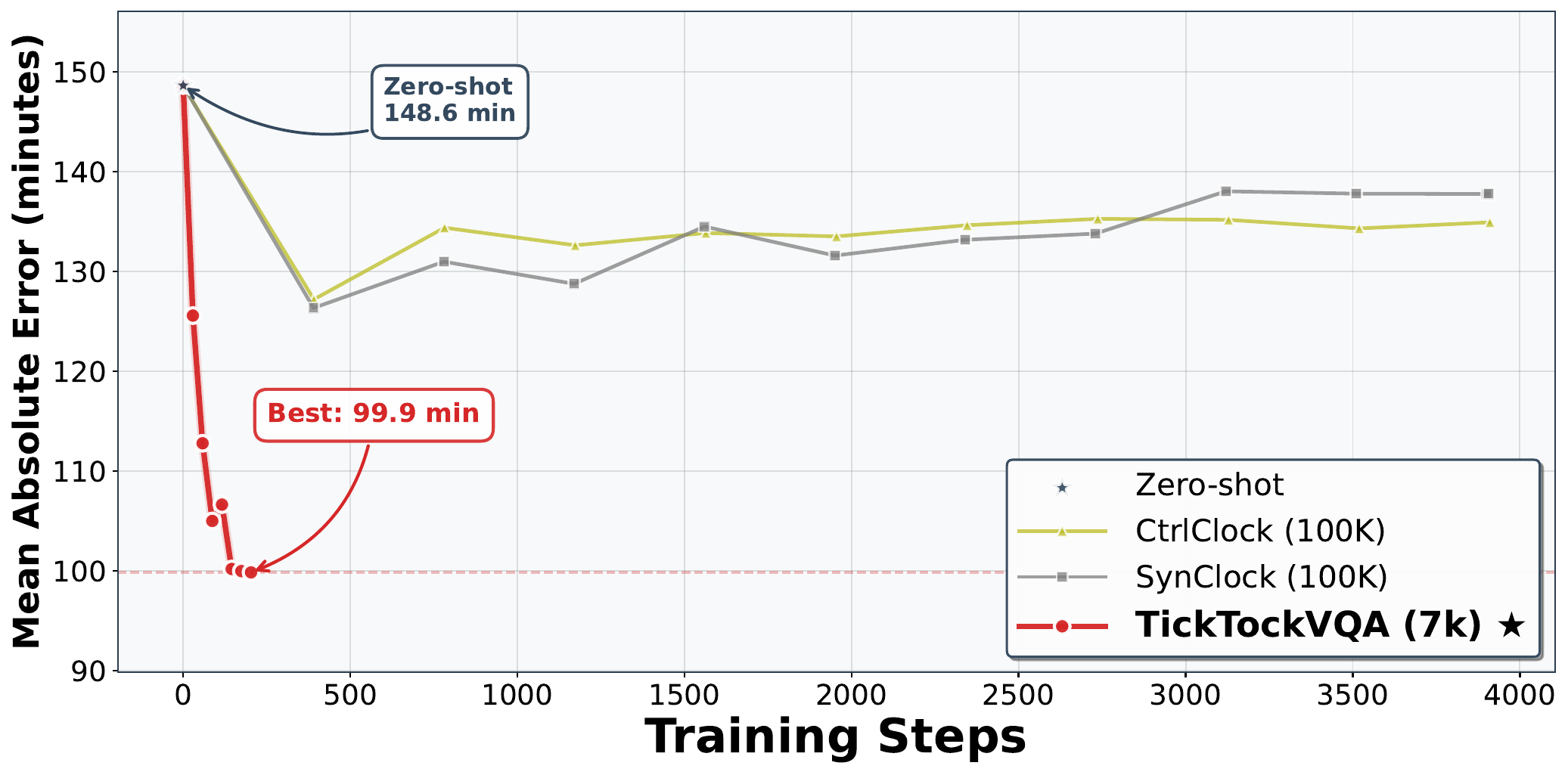}
    \caption{
    \textbf{Impact of training data quality on Qwen2.5-VL-7B performance.}
    We compare Qwen2.5-VL-7B trained on three datasets: TickTockVQA (real-world), SynClock (OpenCV-based synthetic), and CtrlClock (diffusion-generated synthetic). Training on TickTockVQA achieves the best performance with 99.9 minutes MAE.}
    \vspace{-1.7em}
    \label{fig:dataset_comparison}
\end{figure}

\begin{figure*}[t]
    \centering
    \includegraphics[width=1.0\linewidth]{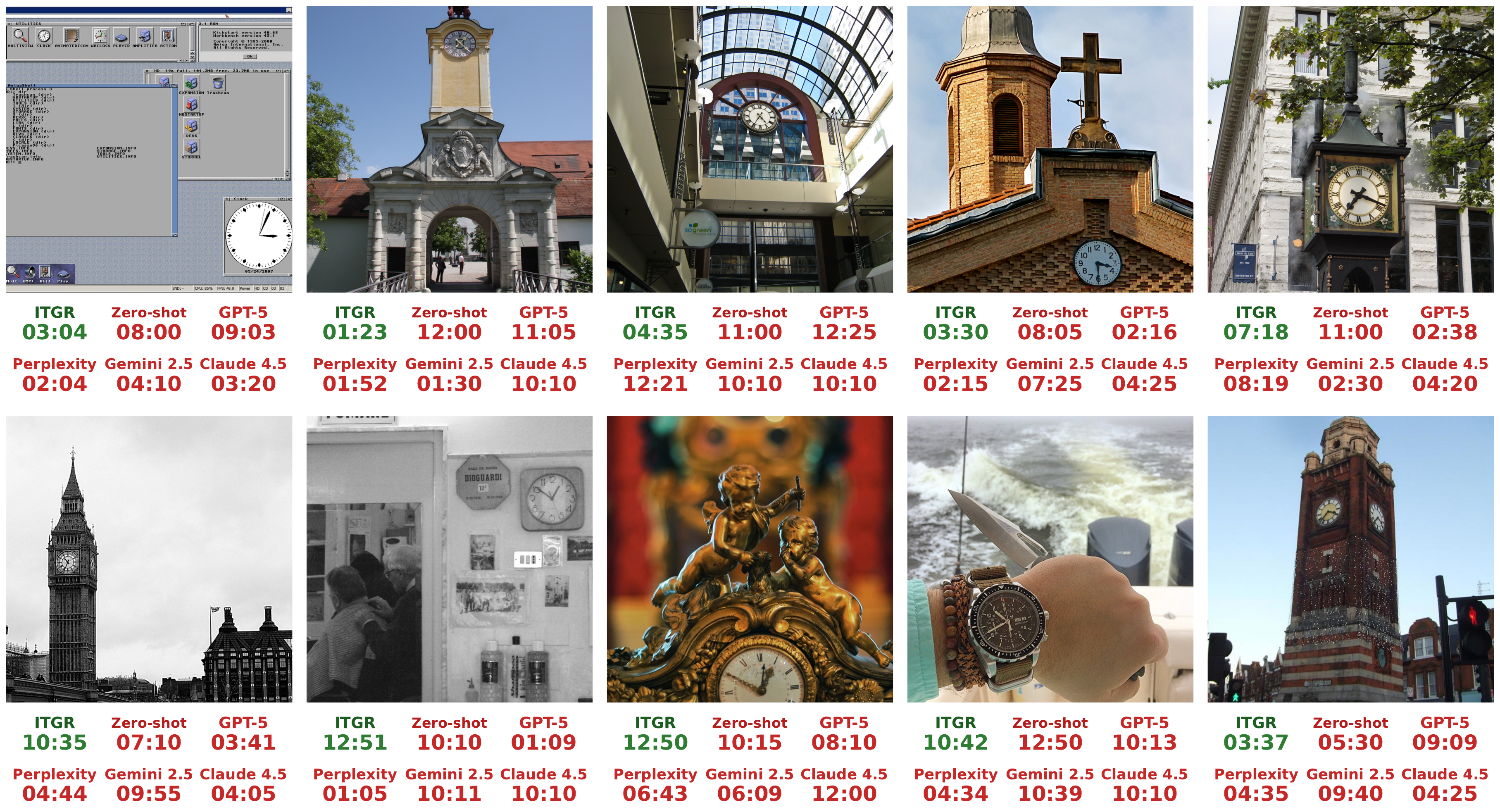}
    \caption{
    Comparison of model predictions on the clock reading task. 
    Our model, It's Time To Get It Right (ITGR), correctly identifies the time, 
    while other large multimodal models (Llama-3.2-11B Zero-shot, GPT-5, Claude Sonnet 4.5, Gemini-2.5 Pro, and Perplexity Pro) produce incorrect results.
    }
    \vspace{-1.5em}
    \label{fig:qualitative_results}
\end{figure*}
Analog clocks are ubiquitous across various contexts, including wall-mounted clocks and tower clocks, and they appear in a wide range of visual styles.
In such settings, clocks exhibit substantial visual variability due to changes in lighting conditions, perspective distortion, and occlusion.
Recent studies demonstrate that even leading multimodal models achieve less than 10\% accuracy on realistic analog clock benchmarks, despite their strong performance on various complex tasks.
We analyze this performance gap and attribute it to two primary factors.
First, there is a lack of large-scale, high-quality datasets specifically curated for analog clock reading in real-world scenes~\cite{saxena2025lost, it's_about_time, safar2025clockbench}.
Publicly available clock images are often biased toward stylized synthetic data or fixed times such as \textit{10:10}~\cite{fu2025have}, which can hinder models' ability to generalize to clock reading scenarios. 
Second, current models exhibit limited spatial reasoning capacity~\cite{liao2024reasoningpathsreferenceobjects,stogiannidis2025mind,gholami2025spatialreasoningegocentric} that is required to interpret time on analog clocks. 
In particular, they struggle with assigning the correct semantic roles to visually similar components, most notably confusing the hour and minute hands.
To address these limitations, we present \textbf{TickTockVQA}, a human-annotated dataset of approximately 12k images collected from real-world scenes.
In contrast to synthetic or stylized datasets, TickTockVQA captures the complexities of real environments where clocks often appear.
In addition, our dataset captures real-world inconsistencies, including variations in clock numbering and clock-hand shapes that differ widely across analog clock designs.
For each image, we provide explicit annotation of the hour, minute, and AM/PM indicators, allowing models to learn a more precise understanding of time grounded in real-world scenarios.
This design choice enables models to move beyond the biases of highly stylized datasets and encourages VLMs to develop robust clock-reading capabilities across naturally diverse contexts as demonstrated in Figure~\ref{fig:dataset_comparison}.

Building on the TickTockVQA dataset, we further apply the direct preference optimization (DPO)~\cite{rafailov2023direct} framework. We present \textbf{Swap-DPO} to fine-tune the model to explicitly align its preferences toward correct interpretations of the hour and minute hands.
By incorporating these diverse cases into our dataset and aligning the model's reasoning with Swap-DPO, we demonstrate that the model not only learns to reliably distinguish the hour hand from the minute hand, but also develops the ability to correctly interpret numbering across different styles.
Overall, our study reveals that combining real-world scene data with Swap-DPO fine-tuning significantly enhances a model's proficiency in reading analog clocks, improving both accuracy and robustness to environmental complexity.
Using Llama-3.2-11B~\cite{dubey2024llama}, we achieve a full-time accuracy of 46.22\% on TickTockVQA, representing an improvement of 44.81 percentage points (pp) over the zero-shot baseline.
As illustrated in Figure~\ref{fig:qualitative_results}, our fine-tuned model correctly reads the time in challenging real-world scenarios where strong proprietary and open-source models fail.
These findings further establish analog clock reading as a principled testbed for advancing spatiotemporal reasoning, which opens a new direction for developing more reliable multimodal systems.
\begin{table}[!t]
\centering
\caption{Zero-shot performance comparison on the TickTockVQA test set.}
\label{tab:zero_shot_summary}
\resizebox{\columnwidth}{!}{
\begin{tabular}{l c c c c}
\toprule
\textbf{Model} & \textbf{Hour Acc} & \textbf{Min. Acc} & \textbf{Full Time Acc} & \textbf{MAE$\downarrow$} \\
\midrule
SpatialVLM-3B\cite{spatialvlm} & 12.51 & 6.44 & 1.05 & 161.68 \\
Llama-3.2-11B~\cite{dubey2024llama} & 11.51 & 8.58 & 1.43 & 156.96 \\
Gemma3-12B\cite{gemma3} & 13.00 & 10.62 & 2.12 & 156.49 \\
InternVL3-8B\cite{internvl3} & 13.59 & 9.08 & 2.20 & 159.92 \\
Qwen2.5-VL-7B~\cite{qwen2.5vl} & 17.65 & 22.44 & 6.04 & 148.62 \\
It's About Time\cite{it's_about_time} & 28.95 & 25.00 & 18.54 & 135.15 \\
\bottomrule
\vspace{-1em}
\end{tabular}%
}
\end{table}

\vspace{-2em}

\section{Related Work}\label{sec:2}

\subsection{Visual Question Answering} 
Visual question answering (VQA) has become a central component in both evaluating and improving the capabilities of modern vision-language models. 
While benchmarks such as TextVQA~\cite{textvqa}, ChartQA~\cite{chartqa}, DocVQA~\cite{docvqa}, OK-VQA~\cite{okvqa} and VQAv2~\cite{vqav2} are widely used for performance measurement, numerous studies have shown that these datasets also serve as powerful training signals that substantially enhance models’ visual recognition, multimodal alignment, and reasoning abilities.
VQA tasks require models to integrate fine-grained visual perception with contextual and semantic understanding, enabling strong generalization across novel scenarios.
For instance, the MMMU benchmark~\cite{mmmu} evaluates models across six broad disciplines—including science, engineering, medicine, and the humanities—demanding expert-level domain knowledge and multi-step reasoning.
Recent large-scale VLMs consistently report that performance gains on such VQA benchmarks strongly correlate with improvements in downstream multimodal tasks~\cite{li2025survey}.
\subsection{Clock Reading VQA: Prior Datasets and Challenges}
Collecting reliable clock data for training is challenging, which constrains models' ability to accurately interpret time from analog clock images.
Yang \textit{et al.}~\cite{it's_about_time} address this issue by generating synthetic clock datasets, which improve model performance on time-reading tasks.
However, such datasets are limited in scale and suffer from low fidelity, making them less representative of real-world scenes.
In parallel, Saxena \textit{et al.}~\cite{saxena2025lost} investigate multimodal large language models' understanding of time and date, demonstrating that most models perform poorly on both clock-reading and calendar-date tasks.
These works highlight that time reading is a fundamental skill for spatiotemporal reasoning, a capability closely tied to computer vision and contextual understanding.

\subsection{Spatial Reasoning}
Understanding spatial information is one of the key challenges for recent VLMs.
Although models have advanced in object recognition and semantic reasoning, their ability to perform fine-grained spatial reasoning remains limited. Recent studies have introduced large-scale datasets and training frameworks aimed at enhancing spatial understanding in VLMs~\cite{stogiannidis2025mind, liu2025mirage, jia2025omnispatial,li2024superclevr}.
Cheng \textit{et al.}~\cite{cheng2024spatialrgpt} 
introduce a depth information plug-in module that enables VLMs to develop a more accurate understanding of spatial arrangement.
Similarly, Chen \textit{et al.}~\cite{spatialvlm} propose a large-scale 3D spatial data generation framework that improves VLMs’ ability to reason about spatial relationships in real-world images.
Furthermore, researchers have identified limitations in current VLMs' fine-tuning approaches for spatial reasoning tasks, including the over-reliance on pre-annotated instruction data~\cite{qiu2025spatialpreference} and the biases in synthetic preference annotations~\cite{vapr}. These issues restrict models' spatial reasoning capabilities and motivate the use of preference-based alignment methods such as DPO to more reliably correct such deficiencies.

\begin{table*}[!t]
    \small
    \centering
    \caption{Breakdown of \textbf{TickTockVQA} instances by clock type. The
    Environment and Transformation columns are mutually exclusive, whereas the Design 
    columns may contain multiple labels.}
    \label{tab:diversity}
    \begin{tabular}{l
    @{\hskip 0.4em}c@{\hskip 0.4em}  
    @{\hskip 0.4em}c@{\hskip 0.4em}  
    @{\hskip 0.4em}c@{\hskip 0.4em}  
    @{\hskip 0.4em}c@{\hskip 0.4em}  
    @{\hskip 0.4em}c@{\hskip 0.4em}  
    @{\hskip 0.4em}c@{\hskip 0.4em}  
    @{\hskip 0.4em}c@{\hskip 0.4em}  
    @{\hskip 0.4em}c@{\hskip 0.4em}  
    @{\hskip 0.4em}c@{\hskip 0.4em}  
    @{\hskip 0.4em}c@{\hskip 0.4em}} 
    \toprule
    \multirow{2}{*}{\textbf{Clock Type}} &
    \multicolumn{3}{c}{\textbf{Environment}} &
    \multicolumn{3}{c}{\textbf{Transformation}} &
    \multicolumn{3}{c}{\textbf{Design (multi-label)}} &
    \multirow{2}{*}{\textbf{Total}} \\
    \cmidrule(lr){2-4} \cmidrule(lr){5-7} \cmidrule(lr){8-10}
    {} & Indoor & Outdoor & Unknown & Normal & Flipped & Partial & Arabic & Roman & No numerals & {} \\
    \midrule
    Wall clocks         & 2107 & 1622 &  317 & 3887 & 10 & 149 & 1953 & 1465 &  671 &  4046 \\
    Tower clocks        &    3 & 1905 &    5 & 1907 &  2 &   4 &  112 &  925 &  881 &  1913 \\
    Wristwatches        &  495 &  214 &  529 & 1214 &  1 &  23 &  795 &   95 &  363 &  1238 \\
    Post clocks         &   62 &  931 &    6 &  992 &  1 &   6 &  321 &  512 &  171 &   999 \\
    Alarm/Desk clocks   & 1235 &   53 &  213 & 1398 & - & 103 &  998 &  282 &  241 &  1501 \\
    Graphic/Illustrated &   48 &   19 &  826 &  845 & - &  48 &  444 &  163 &  301 &   893 \\
    ETC                 &  759 &  575 &  559 & 1817 &  4 &  72 &  863 &  806 &  300 &  1893 \\
    \midrule
    Total               & 4709 & 5319 & 2455 & 12060 & 18 & 405 & 5486 & 4248 & 2928 & 12483 \\
    \bottomrule
    \end{tabular}

    \vspace{0.25em}
    \begin{minipage}{0.96\textwidth}\footnotesize
    \textbf{Notes.} 
    The Environment (Indoor/Outdoor/Unknown) and Transformation (Normal/Flipped/Partial) labels are mutually exclusive within their respective categories, whereas Design is multi-label.
    \vspace{-1.2em}
    \end{minipage}
\end{table*}

\begin{figure}[t]
    \centering

    \begin{subfigure}[b]{0.23\textwidth }
        \includegraphics[width=\columnwidth]{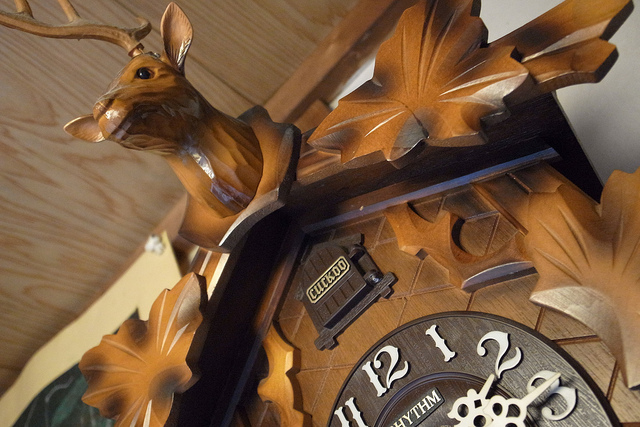}
        \caption{Cropped clock}
        \label{fig:figure1a}
    \end{subfigure}
    \hfill
    \begin{subfigure}[b]{0.23\textwidth }
        \includegraphics[width=\columnwidth]{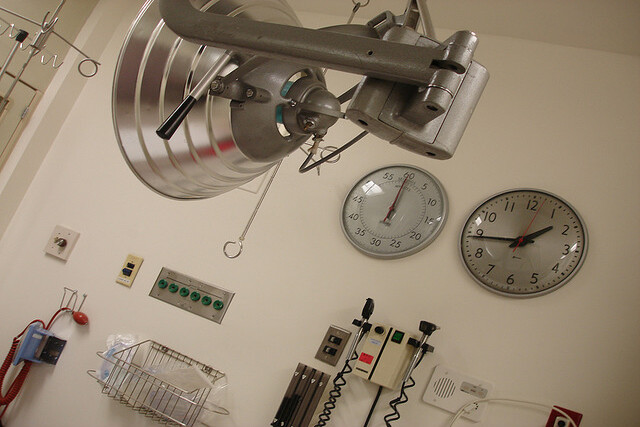}
        \caption{Clock-like object}
        \label{fig:figure1b}
    \end{subfigure}
    
    \begin{subfigure}[b]{0.23\textwidth }
        \includegraphics[width=\columnwidth]{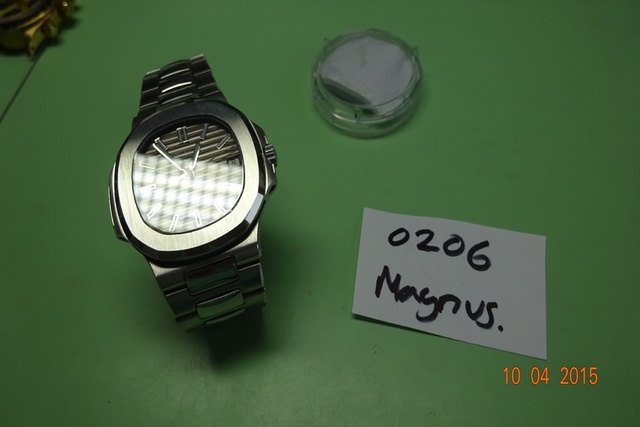}
        \caption{Illumination changes}
        \label{fig:figure2a}
    \end{subfigure}
    \hfill
    \begin{subfigure}[b]{0.23\textwidth }
        \includegraphics[width=\columnwidth]{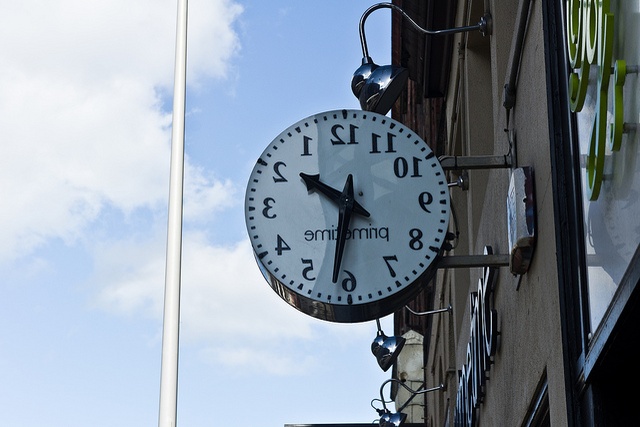}
        \caption{Horizontally flipped clock}
        \label{fig:figure2b}
    \end{subfigure}

    \vspace{-0.5em}
    \caption{Examples of challenging visual variations in the \textbf{TickTockVQA} test set: (a) cropped clock, (b) clock-like object, (c) illumination changes, and (d) horizontally flipped clock. 
  The figure highlights diverse transformations and ambiguities that models must handle for robust clock understanding.}
  \label{fig:ticktock_variations}
  \vspace{-1.5em}
\end{figure}

\section{TickTockVQA: A Real-World Benchmark}\label{sec:3}

\subsection{Collection Pipeline}
We curated approximately 12k real-world clock images from diverse sources, including COCO~\cite{lin2014microsoft}, SBU Captions~\cite{ordonez2011im2text}, Visual Genome (VG)~\cite{krishna2017visual}, ImageNet~\cite{deng2009imagenet}, Open Images (OID)~\cite{kuznetsova2020open}, Conceptual Captions 12M (CC12M)~\cite{changpinyo2021conceptual}, and movie frames (e.g., the \textit{Clock Movie})~\cite{it's_about_time}. 
For COCO, VG, ImageNet~\cite{deng2009imagenet}, and OID, we leveraged existing object annotations to directly extract images containing the clock class, followed by manual verification. 
To further ensure data quality, we detected and removed exact and near-duplicate images between VG and COCO by computing SHA-1 and perceptual hashes (pHash, wHash). For SBU and CC12M, which are web-crawled caption datasets, we filtered candidate images based on captions containing keywords such as \textit{clock} or \textit{watch}.
To improve precision, we excluded irrelevant matches (e.g., watching) and removed digital clocks, focusing solely on analog instances.
All candidate images then underwent a second round of manual inspection to ensure they were visually interpretable.
We addressed the over-representation of canonical times such as 10:10, which frequently appear in stock photos or product advertisements.
We retained only a subset of such instances, ensuring a more balanced temporal distribution across the dataset.

\subsection{Annotation Protocol}
The authors manually annotated each clock image with the corresponding hour and minute.
When the scene context permitted, an additional AM/PM tag was assigned.
In ambiguous cases, such as sky scenes or indoor lighting where both \textit{6 AM} and \textit{6 PM} were plausible, no AM/PM label was provided.
When commonsense reasoning offered sufficient cues, the images were consistently labeled AM or PM.
To ensure annotation quality, every instance was independently labeled by at least two authors, and any disagreements were resolved by consensus.

\subsection{Dataset Diversity and Statistics}
Table~\ref{tab:diversity} provides a detailed breakdown of these factors across clock categories. As shown, TickTockVQA distributes clock types broadly across environments (e.g., wall clocks indoors vs. tower clocks outdoors) and captures substantial diversity in both visual transformations and face designs, yielding a comprehensive benchmark for robust analog-clock understanding. To further illustrate the visual challenges present in TickTockVQA, 
Figure~\ref{fig:ticktock_variations}
showcases representative examples from the test set.
Such cases demonstrate the complex visual ambiguities that models must resolve, ranging from differentiating true clocks from impostor objects to reasoning under geometric distortions and ambiguous lighting conditions.
In total, TickTockVQA contains 12,483 annotated analog clocks, making it, to our knowledge, the largest and most diverse in-the-wild benchmark for analog clock understanding.
TickTockVQA consists of 12,483 images, comprising 7,236 training and 5,247 test samples.
For fair comparison with prior work, we adopt the same test sources as those used in It’s About Time~\cite{it's_about_time}, ensuring cross-dataset compatibility while providing a substantially larger and cleaner benchmark for evaluating time-reading capability in the wild.
\vspace{-4.5em}

\vspace{5.0em}
\begin{table*}[!t]
\centering
\footnotesize
\caption{Comprehensive evaluation results for Gemma3-12B, Qwen2.5-VL-7B, and Llama-3.2-11B on the TickTockVQA test set. B and S denote baseline and swap-equivalence evaluation, respectively (Sec~\ref{sec:eval_protocol}.)}
\label{tab:model_ablations}
\setlength{\tabcolsep}{3pt}
\begin{tabular}{l@{\extracolsep{9pt}} cc cc cc cc cc cc}
\toprule
\multirow{2}{*}{\textbf{Model and Training Stages}}
  & \multicolumn{2}{c}{\textbf{Hour Acc}}
  & \multicolumn{2}{c}{\textbf{Minute Acc}}
  & \multicolumn{2}{c}{\textbf{Full Time Acc}}
  & \multicolumn{2}{c}{\textbf{MAE$\downarrow$ (hour)}}
  & \multicolumn{2}{c}{\textbf{MAE$\downarrow$ (minute)}}
  & \multicolumn{2}{c}{\textbf{MAE$\downarrow$ (total)}} \\
\cmidrule(lr){2-3} \cmidrule(lr){4-5} \cmidrule(lr){6-7}
\cmidrule(lr){8-9} \cmidrule(lr){10-11} \cmidrule(lr){12-13}
  & B & S & B & S & B & S & B & S & B & S & B & S \\
\midrule
Gemma3-12B\cite{gemma3}
& 13.00 & 20.75 & 10.62 & 19.36 & 2.12 & 3.05 & 2.61 & 2.59 & 14.43 & 14.32 & 156.49 & 155.10 \\
\quad w/ TickTockVQA
& \underline{46.37} & \underline{55.04} & \underline{54.39} & \underline{65.68} & \underline{34.21} & \underline{37.11} & \textbf{1.35} & \textbf{1.27} & \underline{6.68} & \underline{6.34} & \textbf{81.66} & \textbf{77.26} \\
\quad w/ Swap-DPO
& \textbf{46.67} & \textbf{56.01} & \textbf{57.54} & \textbf{67.64} & \textbf{35.32} & \textbf{37.89} & \underline{1.36} & \underline{1.29} & \textbf{6.16} & \textbf{5.85} & \underline{81.86} & \underline{77.97} \\
\midrule
Qwen2.5-VL-7B~\cite{qwen2.5vl}
& 17.65 & 25.26 & 22.44 & 32.14 & 6.04 & 7.26 & 2.47 & 2.44 & 11.96 & 11.81 & 148.62 & 146.76 \\
\quad w/ SynClock
& 28.11 & 37.51 & 35.68 & 47.17 & 16.12 & 18.28 & 2.15 & 2.09 & 9.60 & 9.35 & 129.51 & 126.34 \\
\quad w/ CtrlClock
& 25.16 & 33.81 & 37.45 & 46.85 & 14.75 & 17.08 & 2.18 & 2.11 & 9.58 & 9.28 & 131.14 & 127.20 \\
\quad w/ TickTockVQA
& \underline{33.18} & \underline{42.94} & \underline{41.51} & \underline{52.70} & \underline{20.34} & \underline{22.76} & \underline{1.73} & \underline{1.66} & \underline{8.53} & \underline{8.19} & \underline{104.31} & \underline{100.20} \\
\quad w/ Swap-DPO
& \textbf{35.39} & \textbf{44.98} & \textbf{46.16} & \textbf{55.29}
& \textbf{23.06} & \textbf{25.08} & \textbf{1.61} & \textbf{1.55}
& \textbf{7.24} & \textbf{7.01} & \textbf{96.42} & \textbf{93.37} \\
\midrule
Llama-3.2-11B\cite{dubey2024llama}
& 11.49 & 18.35 & 8.58 & 16.85 & 1.41 & 2.02 & 2.59 & 2.57 & 14.79 & 14.70 & 156.96 & 155.67 \\
\quad w/ SynClock
& 32.55 & 40.67 & 39.24 & 47.04 & 22.09 & 23.80 & 2.10 & 2.04 & 9.77 & 9.54 & 126.77 & 123.79 \\
\quad w/ CtrlClock
& 33.28 & 40.10 & 40.10 & 50.09 & 23.04 & 25.00 & 2.12 & 2.06 & 9.95 & 9.65 & 128.10 & 124.34 \\
\quad w/ TickTockVQA
& \underline{56.05} & \underline{64.40} & \underline{66.69} & \underline{73.57} & \underline{45.78} & \underline{48.10} & \underline{1.04} & \underline{0.98} & \textbf{4.82} & \textbf{4.56} & \underline{62.52} & \underline{59.22} \\
\quad w/ Swap-DPO
& \textbf{56.41} & \textbf{64.42} & \textbf{66.69} & \textbf{73.62} & \textbf{46.22} & \textbf{48.48} & \textbf{1.03} & \textbf{0.97} & \underline{4.85} & \underline{4.61} & \textbf{61.93} & \textbf{58.79} \\
\bottomrule
\end{tabular}
\end{table*}

\section{Method}\label{sec:4}


\subsection{Base Models}
Our method is applied to multiple state-of-the-art open-source VLMs to demonstrate its generality and scalability.  
Specifically, we adopt Qwen2.5-VL-7B~\cite{qwen2.5vl}, Llama-3.2-11B~\cite{dubey2024llama}, 
and Gemma3-12B~\cite{gemma3} as the base architectures.
%
\begin{algorithm}[t]
\caption{End-to-end pseudo-code of the proposed two-stage VLM fine-tuning pipeline}
\label{alg:main_summary}
\begin{algorithmic}[1]
\Require Pre-trained VLM $\mathcal{M}_0$, training data $\mathcal{D}_{\text{train}} = \{(x, y_{\text{gt}})\}$
\Ensure DPO-tuned model $\mathcal{M}_{\text{DPO}}$

\State $\mathcal{M}_{\text{SFT}} \gets \textsc{LoRA-SFT}(\mathcal{M}_0, \mathcal{D}_{\text{train}})$  \Comment{Adapt model to clock domain}
\State Initialize $\mathcal{D}_{\text{pref}} \gets \emptyset$

\For{each $(x, y_{\text{gt}})$ in $\mathcal{D}_{\text{train}}$}
  \State $\hat{y} \gets \mathcal{M}_{\text{SFT}}(x)$
  \If{not $\textsc{IsCorrect}(\hat{y}, y_{\text{gt}})$}
      \State $y_{\text{rej}} \gets \hat{y}$
  \Else
      \State $y_{\text{rej}} \gets \textsc{SwapHands}(y_{\text{gt}})$
  \EndIf
  \State Add $(x, y_{w}=y_{\text{gt}}, y_{l}=y_{\text{rej}})$ to $\mathcal{D}_{\text{pref}}$
\EndFor

\State $\mathcal{M}_{\text{policy}} \gets \textsc{MergeLoRA}(\mathcal{M}_0, \mathcal{M}_{\text{SFT}})$
\State $\mathcal{M}_{\text{DPO}} \gets \textsc{DPO-Train}(\pi_{\theta}=\mathcal{M}_{\text{policy}}, \pi_{\text{ref}}=\mathcal{M}_{\text{policy}}, \mathcal{D}_{\text{pref}})$
\State \Return $\mathcal{M}_{\text{DPO}}$
\end{algorithmic}
\end{algorithm}
\vspace{-1.0em}
\subsection{Fine-tuning Strategy}
To enhance the clock-reading capabilities of the base VLM, we propose a two-stage fine-tuning process.
First, we perform supervised fine-tuning (SFT) using low-rank adaptation (LoRA)~\cite{hu2022lora} to train the model on the fundamental task of identifying clock hands in analog clock images for accurate time reading.
Although SFT adapts the model to the clock domain, it provides no mechanism to enforce consistent semantic roles for the hour and minute hands. Consequently, the model often interprets the shorter and longer hands interchangeably in challenging configurations.
To address this specific hand-swapping confusion, we apply a variant of DPO~\cite{rafailov2023direct}, which we term \textbf{Swap-DPO}.
The core idea of Swap-DPO is to explicitly teach the model to prefer the correct time, denoted as $y_w$, over a hard negative sample $y_l$ generated by swapping the roles of the hour and minute hands. We synthesize this hard negative sample by geometrically reinterpreting the hands' angular positions.
Let $\theta_h = 30h + \tfrac{m}{2}$ and $\theta_m = 6m$ denote the angular positions of the hour and minute hands in degrees, respectively.
The swapped time $h_{\text{new}}$ and $m_{\text{new}}$, which form the basis of the rejected response $y_l$, are derived as follows:

\[
\begin{aligned}
    h_{\text{new}} &= \Big\lfloor \frac{\theta_m}{30} \Big\rfloor, &
    m_{\text{new}} = \Big( \frac{\theta_h}{6} \Big) \bmod 60
\end{aligned}
\]

where $h \in [0, 11]$ and $m \in [0, 59]$. This formulation yields a geometrically consistent but incorrect time, forcing the model to learn the distinct roles of the hands.
We optimize the model using the standard DPO objective to encourage the policy to prefer the correct time annotation over the constructed hard negative:
{\setlength{\abovedisplayskip}{4pt}
 \setlength{\belowdisplayskip}{4pt}
\begin{equation}
\begin{split}
\mathcal{L}_{\text{DPO}}(\pi_\theta; \pi_{\text{ref}}) 
= -\mathbb{E}_{(x, y_w, y_l) \sim \mathcal{D}} \Big[
\log \sigma \Big(
\beta \log \tfrac{\pi_\theta(y_w|x)}{\pi_{\text{ref}}(y_w|x)}
\\
\qquad\qquad
- \beta \log \tfrac{\pi_\theta(y_l|x)}{\pi_{\text{ref}}(y_l|x)}
\Big) \Big].
\end{split}
\end{equation}
}
where $\pi_{\text{ref}}$ is the frozen reference model. 

Our contribution lies not in the DPO loss itself, but in constructing clock-specific preference pairs through our Swap-DPO formulation.
To build the preference dataset $\mathcal{D}$, we perform inference on the training set using the SFT model. 
For each sample, $y_w$ is the ground-truth time, and the rejected response $y_l$ is defined as follows:
\begin{enumerate}
    \item If the SFT model produces a clearly incorrect prediction, we directly use it as $y_l$.
    \item If the prediction is near-correct, we generate a hard negative by applying the Swap-DPO transformation to the ground-truth time.
\end{enumerate}
This strategy yields approximately 7k preference pairs, enabling the model to focus on clock-specific ambiguities—particularly hour–minute hand confusion. This entire process is summarized in Algorithm~\ref{alg:main_summary}.

\section{Experiments}\label{sec:5}
Unless otherwise stated, all reported results and figures use Llama-3.2-11B~\cite{dubey2024llama} as the backbone.
\vspace{-0.2em}
\begin{figure*}[t]
    \centering
    \includegraphics[width=1.0\linewidth]{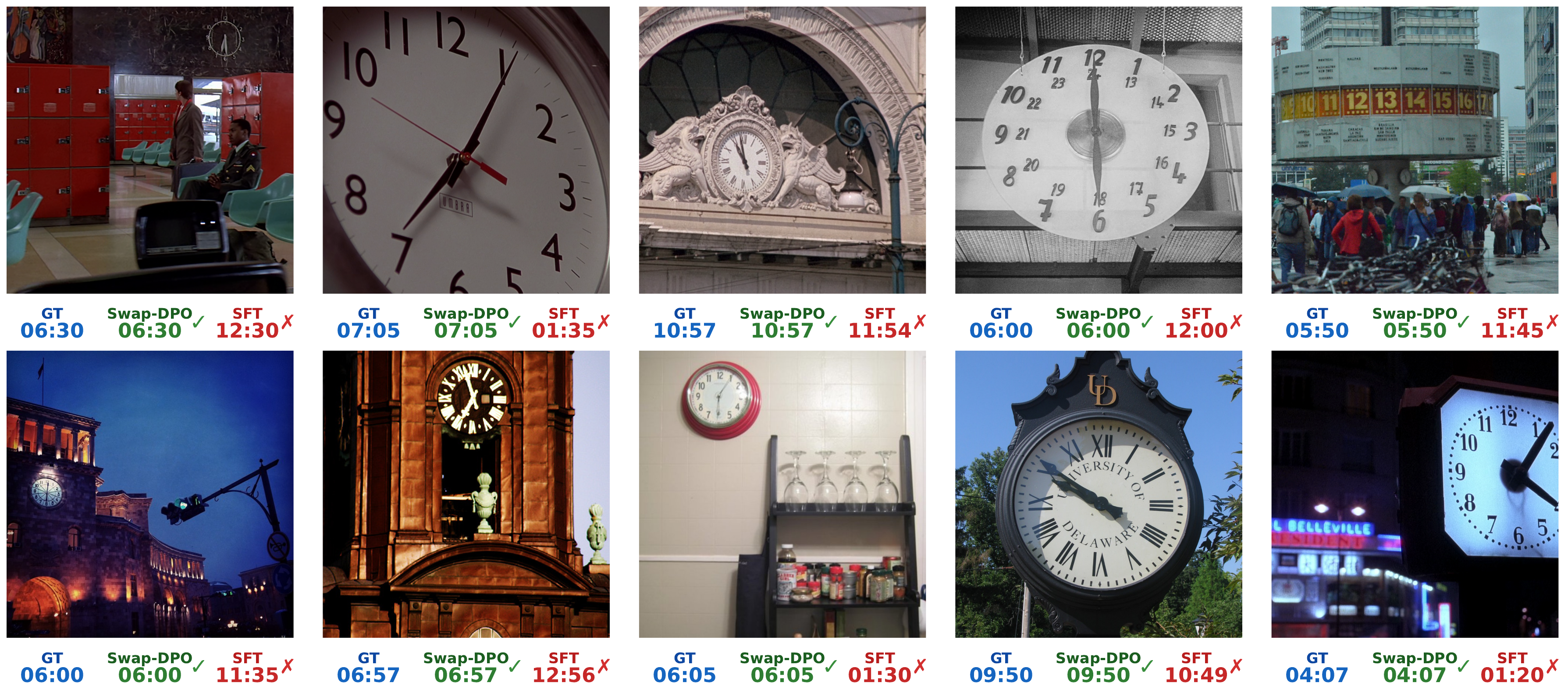}
    \caption{
    \textbf{Qualitative examples of hand-swap error correction by Swap-DPO.} 
    SFT incorrectly swaps the hour and minute hands, whereas Swap-DPO successfully corrects this systematic error pattern.
    }
    \vspace{-1.5em}
    \label{fig:handswap_qualitative_results}
\end{figure*}

\subsection{Implementation Details} 
In the SFT stage, we adapt the model to the clock domain using LoRA~\cite{hu2022lora}. 
It is applied to all linear layers in both the vision tower and the language model, excluding the language model head and embedding layers. 
During SFT, the base model parameters remain frozen while only the adapter weights are updated. 
For the Swap-DPO stage, we continue applying LoRA-based fine-tuning to both the vision and language components of the model. 
We additionally adopt a differential learning-rate scheme to reflect the varying sensitivity of different modules during preference optimization. All experiments are conducted on a cluster of 8 NVIDIA A6000 GPUs, taking approximately 8 hours per full training run. 
%
\subsection{Evaluation Protocol}
\label{sec:eval_protocol}
We evaluate all models on the TickTockVQA test set.
To comprehensively assess clock-reading performance, we use a multifaceted evaluation protocol.
First, we measure overall accuracy using hour accuracy, minute accuracy (with a ±2 minute tolerance), and full-time accuracy.
While accuracy metrics measure success, they fail to distinguish between minor inaccuracies and severe failure cases. Therefore, we report mean absolute error (MAE), which serves as a complementary metric capturing the magnitude of the temporal deviation. 
%
%
Second, to specifically isolate and diagnose the hand-swapping problem, we evaluate models under two settings: Baseline (B) and Swap-equivalence (S). 
In the Swap-equivalence setting, a prediction is considered correct even if the hour and minute hands are swapped. 
The resulting gap between (B) and (S) scores provides a direct, quantitative measure of the model's confusion between the hands.
A primary goal of our fine-tuning strategy is to minimize this gap, demonstrating that the model can not only locate the hands but also correctly distinguish their roles.
We compare our fine-tuned models Qwen2.5-VL-7B~\cite{qwen2.5vl}, Llama-3.2-11B~\cite{dubey2024llama}, and Gemma3-12B~\cite{gemma3} with the zero-shot performance of representative approaches, including SpatialVLM~\cite{spatialvlm}, InternVL3-8B~\cite{internvl3} and It’s About Time~\cite{it's_about_time}.

\subsection{Main Results and Analysis}
We organize our main results around three questions:  
(1) How severe is the hand confusion problem in existing VLMs?  
(2) How effectively does SFT on TickTockVQA improve clock understanding?  
(3) Can the proposed Swap-DPO strategy correct this specific spatial reasoning error?
\vspace{-1.5em}

\begin{figure}[t]
    \centering
    \includegraphics[width=1.0\linewidth]{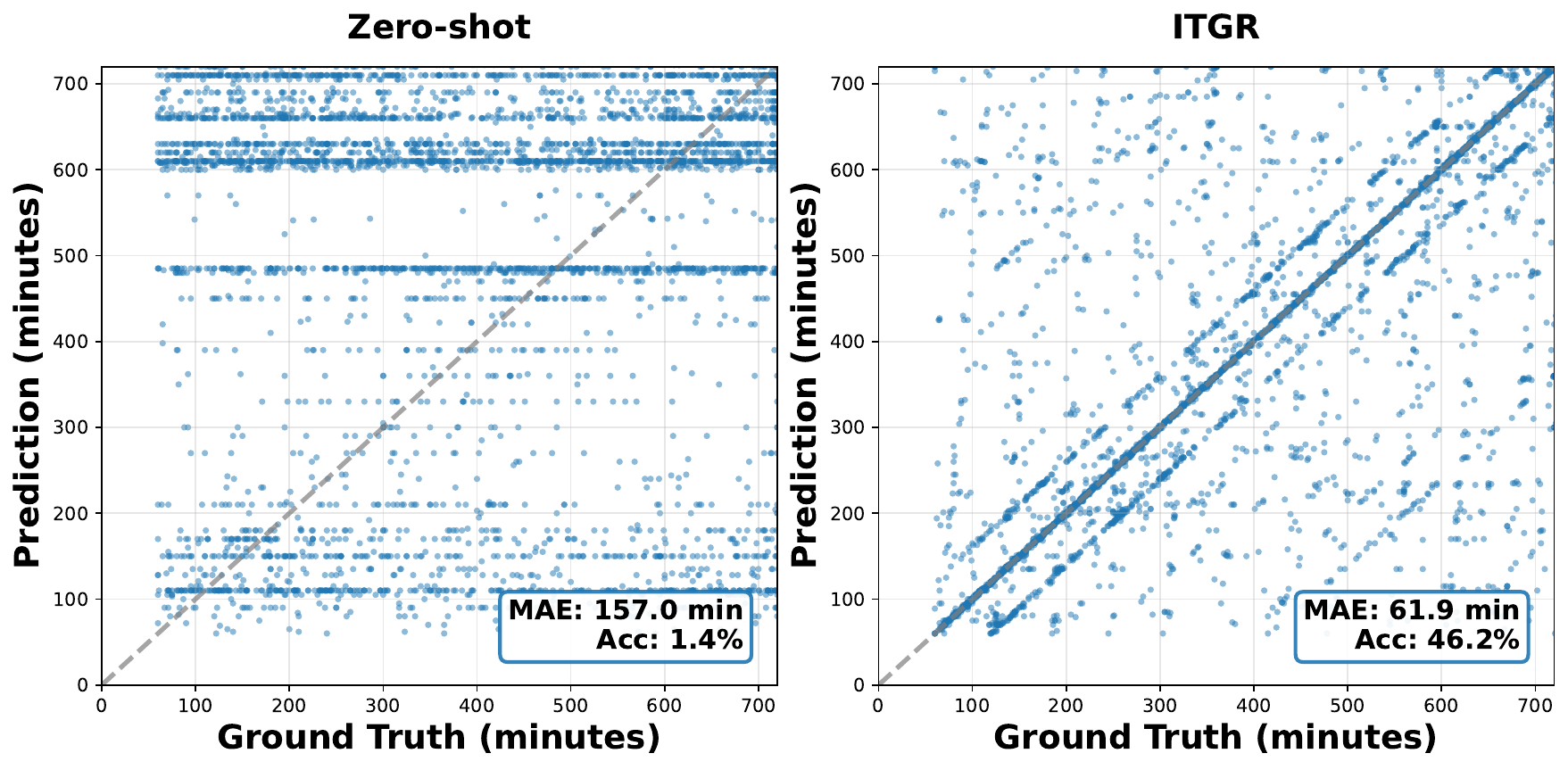}
    \caption{
    \textbf{Quantitative comparison of clock reading accuracy.}
    Each plot visualizes the relationship between ground truth (x-axis) and model-predicted time (y-axis) in minutes. The gray dashed line ($y{=}x$) indicates perfect predictions. 
    Left: Zero-shot baseline. Right: Our ITGR model with the Swap-DPO framework.
    }
    \label{fig:scatter_zeroshot_vs_dpo}
\end{figure}

\paragraph{Baseline Performance and Hand Confusion.} As shown in Table~\ref{tab:zero_shot_summary}, zero-shot predictions exhibit noticeable bias and remain very far from usable performance, with accuracies near random-guessing levels.
This systematic failure is clearly visible in the zero-shot scatter plot in Figure~\ref{fig:scatter_zeroshot_vs_dpo} (left), where predictions form a distinctive off-diagonal cluster—indicating that predictions cluster around a specific biased pattern rather than reflecting random noise.
For example, Llama-3.2-11B~\cite{dubey2024llama} achieves only 1.41\% full-time accuracy with a substantial gap between Baseline and Swap-equivalent minute accuracies (8.58\% vs. 16.85\%), further confirming this structural ambiguity.

\vspace{-1.0em}
\paragraph{Supervised Fine-tuning on TickTockVQA.}
As shown in Table~\ref{tab:model_ablations}, 
SFT on TickTockVQA demonstrates substantial improvements across all VLMs.
Starting from zero-shot baselines, Llama-3.2-11B achieves the most dramatic improvement, with full-time accuracy increasing from 1.41\% to \textbf{45.78\%}—a \textbf{44.37} pp gain.
Similarly, Gemma3-12B advances from 2.12\% to \textbf{34.21\%}, while Qwen2.5-VL-7B improves from 6.04\% to \textbf{20.34\%}.
These consistent gains across diverse model architectures empirically validate that TickTockVQA substantially improves the fine-grained spatiotemporal reasoning capabilities of models for analog clock reading.
However, the persistent gap between (B) and (S) metrics, averaging 2.54\% across SFT models, indicates that spatial ambiguity between hour and minute hands remains unresolved by supervised learning alone.

\vspace{-1.0em}
\paragraph{Correcting Hand Confusion with Swap-DPO.}
While SFT provides a strong performance baseline, it fails to resolve the core ambiguity between hands.
The SFT-trained Qwen2.5-VL-7B model still exhibits a significant 2.42\% hand-swap gap.
To target this specific failure mode, we apply Swap-DPO and demonstrate its effectiveness.
Swap-DPO forces the model to distinguish between the hands' semantic roles.
It narrows the hand-swap gap from 2.42\% to 2.02\%.
By resolving this critical spatial confusion, Swap-DPO also yields consistent gains in overall performance, boosting the final full-time accuracy to 23.06\% and reducing the total MAE from 104.31 to \textbf{96.42} minutes.
We visualize representative hand-swap corrections in Figure~\ref{fig:handswap_qualitative_results}.

\vspace{-1.0em}

\begin{figure}[t]
    \centering
    \includegraphics[width=1.0\linewidth]{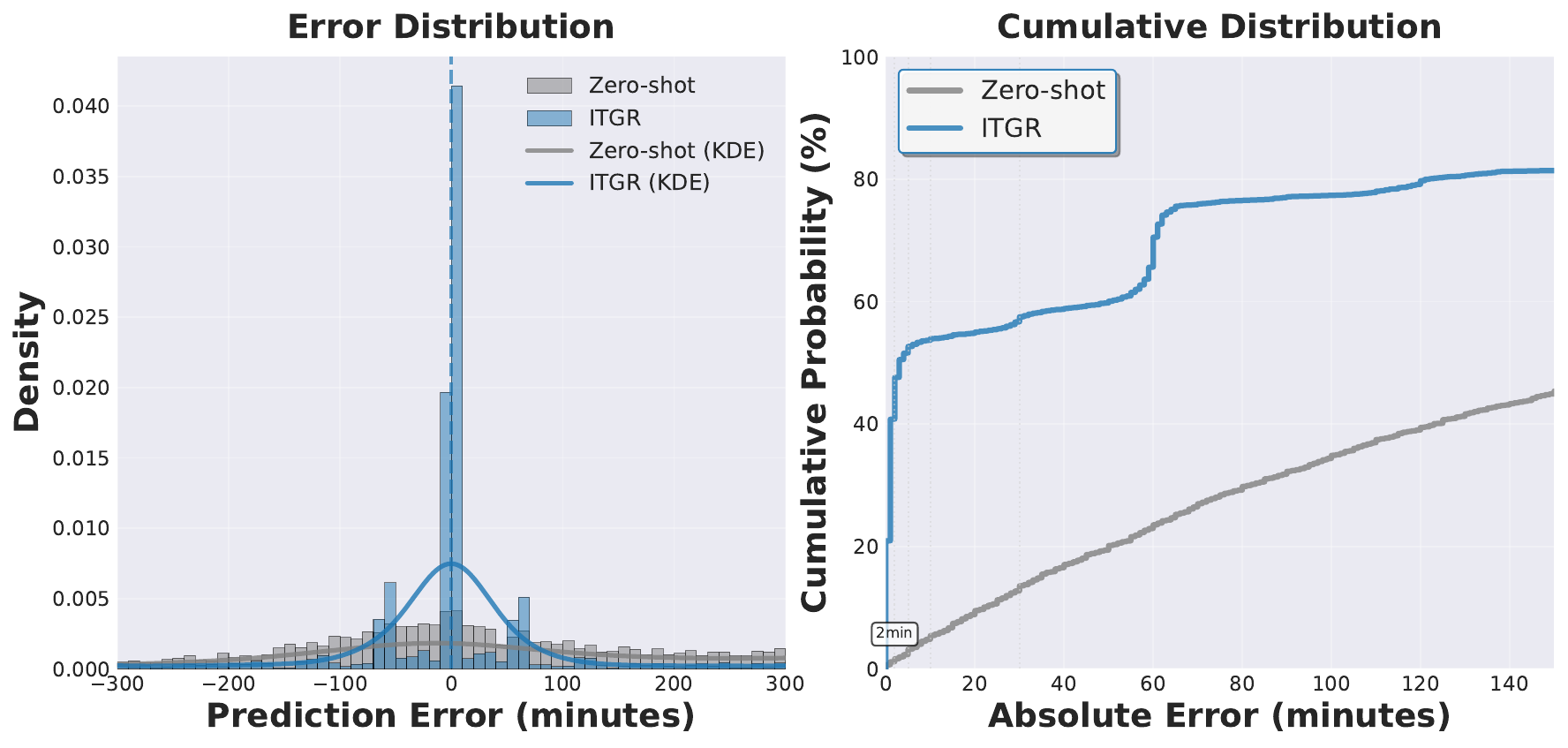}
    \caption{
    \textbf{Statistical analysis of time reading errors.}
    Left: Distribution of prediction errors (in minutes) with histogram and kernel density estimation (KDE). 
    ITGR reduces the heavy tail of large errors.
    Right: Cumulative probability of samples within a given absolute error threshold. Compares Zero-shot baseline vs. ITGR model.
    }
    \vspace{-1em}
    \label{fig:sft_vs_itgr_distribution}
\end{figure}

\paragraph{End-to-end improvement over zero-shot.}
Figures~\ref{fig:scatter_zeroshot_vs_dpo} and~\ref{fig:sft_vs_itgr_distribution} highlight the end-to-end improvement of the final ITGR model over the zero-shot baseline.
As shown in Figure~\ref{fig:scatter_zeroshot_vs_dpo}, zero-shot predictions form a distinctive off-diagonal cluster, whereas ITGR predictions align tightly with the $y{=}x$ diagonal.
Figure~\ref{fig:sft_vs_itgr_distribution} further characterizes this shift in terms of the full error profile: ITGR produces a much sharper peak around zero error with a substantially reduced heavy tail, indicating fewer high-severity mistakes.
Consistently, the cumulative distribution rises markedly faster, showing that a larger fraction of samples fall within practical absolute-error tolerances.

\subsection{Ablation Studies}

\subsubsection{Synthetic Datasets for Comparative Analysis}
To investigate the role of data realism and diversity, we construct two synthetic datasets that serve as controlled counterparts to our real-world benchmark.
The synthetic datasets are used exclusively for comparison rather than joint training.
The first dataset, SynClock, is a lightly modified version of the OpenCV-based set introduced in It’s About Time~\cite{it's_about_time}.
The second, higher-fidelity dataset, CtrlClock, is generated via a diffusion-based controlled synthesis pipeline.
The generation process for CtrlClock leverages an SDXL~\cite{SDXL,LDM} model guided by ControlNet~\cite{controlnet} through a lightweight Ctrl-Adapter~\cite{ctrl-adapter}.
Each image begins from an OpenCV-rendered clock showing an exact time.
The Ctrl-Adapter pipeline extracts a Canny edge map as a structural condition, which is paired with diverse text prompts describing various visual styles (e.g., minimalist modern, classic vintage, industrial artistic), materials (e.g., polished dark wood, brushed aluminum), and lighting conditions (e.g., dramatic shadows, bright airy aesthetic).
This controlled yet varied process yields photorealistic clocks that maintain temporal precision while exhibiting high stylistic diversity.
This design allows us to isolate the effect of data realism by directly comparing models trained on synthetic datasets against those trained on real-world images under identical fine-tuning settings.

\subsubsection{The Impact of Data Realism on Clock Reading.}
\begin{figure}[t]
    \centering
    \includegraphics[width=\linewidth]{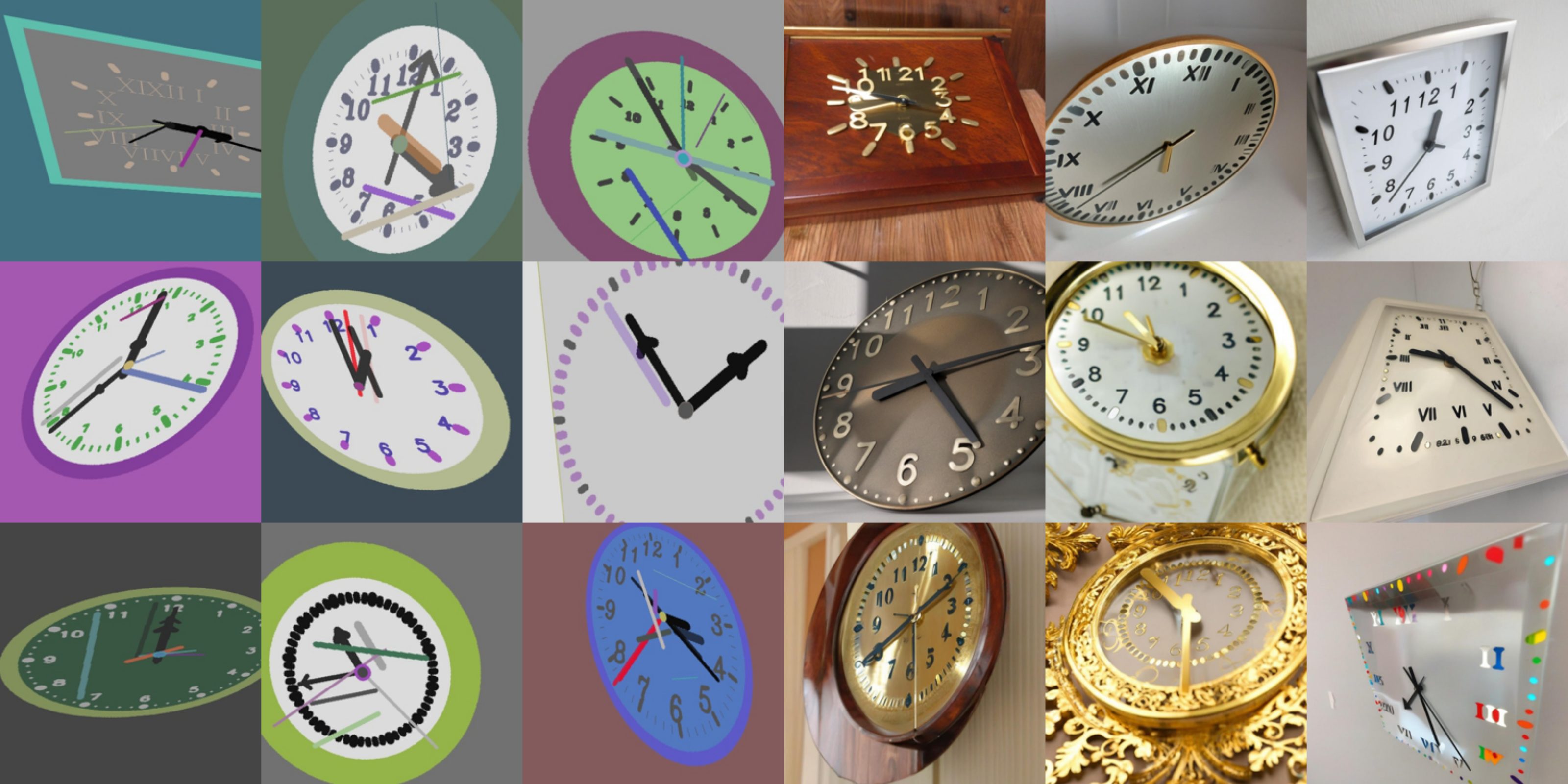}
    \caption{\textbf{Qualitative comparison of synthetic data.} (Left) The SynClock examples, generated using an OpenCV-based approach, exhibit limited realism and flat textures. (Right) The CtrlClock examples, generated through a diffusion-based pipeline, achieve much higher photographic realism and contextual diversity.}
    \vspace{-1.0em}
    \label{fig:syn_ctrl}
\end{figure}

As illustrated in Figure~\ref{fig:dataset_comparison}, the quality and realism of training data have a substantial impact on clock-reading performance.
To isolate and quantify the impact of data realism, we conduct a controlled ablation study. We fine-tune Qwen2.5-VL-7B exclusively on three different datasets—SynClock, CtrlClock, and TickTockVQA—using identical SFT configurations. 
As summarized in Table~\ref{tab:model_ablations}, the results reveal a significant performance gap. Models trained purely on synthetic data achieve limited full-time accuracy. SynClock scores 16.12\% (129.51 total MAE), while the high-fidelity CtrlClock dataset surprisingly achieves a lower score of 14.75\% (131.14 total MAE).
This counter-intuitive finding, where the graphically simpler SynClock outperforms the photorealistic CtrlClock, is noteworthy. 
We hypothesize that this results from an inherent limitation of diffusion-based generative models, 
which often struggle to fully preserve structural conditions during the synthesis process.
While CtrlClock provides superior visual realism (see Figure~\ref{fig:syn_ctrl}), the diffusion pipeline may fail to maintain the absolute spatial fidelity of the clock hands required for precise spatial alignment. This can introduce subtle artifacts or positional jitter—minor deviations that are imperceptible to humans but highly detrimental to a model’s fine-grained spatiotemporal reasoning.
Clock reading is exceptionally sensitive to such minute spatial deviations. In contrast, the less realistic SynClock provides a spatially exact ground truth, offering better results for this specific task.
However, both synthetic approaches are significantly outperformed by TickTockVQA. The exact same architecture trained exclusively on TickTockVQA attains a substantially higher score of 20.34\% with a much lower MAE of 104.31. 
This strongly highlights that the complexity and diversity found in real-world environments are crucial for robust spatiotemporal reasoning. Our findings suggest that simply scaling synthetic data or increasing photorealism is insufficient to capture the nuanced challenges of real-world clock reading.

\begin{table}[!t]
\centering
\caption{Effect of SynClock scale on Qwen2.5-VL-7B under SFT.}
\label{tab:ablation_synclock_scale}
\resizebox{\columnwidth}{!}{
\begin{tabular}{l c c c c}
\toprule
\textbf{Dataset Scale} & \textbf{Hour Acc} & \textbf{Minute Acc} & \textbf{Full Time Acc} & \textbf{MAE$\downarrow$} \\
\midrule
SynClock 12k  & 25.93 & 31.60 & 7.07 & 144.05 \\
SynClock 50k  & 30.19 & 39.74 & 10.58 & 136.65 \\
SynClock 100k & 36.35 & 45.67 & 17.56 & 131.45 \\
SynClock 1M   & 37.51 & 47.17 & 18.28 & 126.34 \\
\bottomrule
\end{tabular}
}
\end{table}

\subsection{Impact of Synthetic Data Scale.}
We further examine how dataset size influences performance by training on SynClock subsets ranging from 12k to 1M samples.
Performance improves as the scale increases but quickly saturates beyond 100k, indicating diminishing returns as illustrated in Table~\ref{tab:ablation_synclock_scale}.
This suggests that simply enlarging synthetic datasets does not necessarily yield richer visual–spatial understanding. Instead, data quality and realism are more crucial.

\subsection{Summary of Findings}
Our experiments lead to the following key observations:
\begin{enumerate}
    \item \textbf{Effectiveness of domain adaptation.} Fine-tuning on the proposed TickTockVQA dataset effectively adapts general-purpose VLMs to the clock-reading domain, enabling robust spatiotemporal reasoning.
    \item \textbf{Impact of preference alignment.} The proposed Swap-DPO strategy substantially mitigates hour–minute hand confusion, demonstrating that targeted preference-based alignment can directly correct fine-grained spatial reasoning errors.
    \item \textbf{Limits of large-scale synthetic data.} While large-scale synthetic datasets improve overall generalization, they still fall short of the representational diversity and contextual realism present in the TickTockVQA images. This suggests that simply scaling synthetic data is insufficient for learning authentic visual–spatial cues as presented in Table \ref{tab:ablation_synclock_scale}.
    \item \textbf{Trade-offs in synthetic generation.} While CtrlClock offers superior photorealism, it may introduce micro-artifacts that harm fine-grained spatial precision. In our experiments, SynClock slightly outperformed CtrlClock, as evidenced in Table~\ref{tab:model_ablations}, suggesting that for this specific task, spatially exact but less realistic data can be more effective than photorealistic data that may introduce subtle noise.
\end{enumerate}

\section{Conclusion}\label{sec:7}
In this work, we examine the long-standing challenge of analog clock reading for vision-language models—a task that requires fine-grained spatiotemporal reasoning.
We introduce \textbf{TickTockVQA}, a 12k real-world benchmark with diverse visual–temporal annotations, and propose \textbf{Swap-DPO}, a targeted preference-alignment method designed to correct hour–minute hand–swapping errors.  
Combined, these contributions substantially improve the full time accuracy of Llama-3.2-11B from \textbf{1.41\%} to \textbf{46.22\%} for analog clock reading in real-world settings.
Despite these advances, performance remains well below human-level~\cite{safar2025clockbench}, highlighting the limitations of current models in fine-grained spatiotemporal reasoning.
%

\section*{Acknowledgments}
We thank Benno Krojer for proofreading the manuscript and providing constructive feedback that improved the clarity of the paper. 
\clearpage
\setcounter{page}{1}
\maketitlesupplementary

\appendix
\renewcommand{\thesection}{\Alph{section}}

\renewcommand{\thefigure}{S\arabic{figure}}
\renewcommand{\thetable}{S\arabic{table}}
\setcounter{figure}{0}
\setcounter{table}{0}


\section{Dataset Analysis}
\label{sec:dataset_analysis}

This section provides detailed statistical analysis of the TickTockVQA dataset, including data source composition, temporal distribution patterns, and filtering strategies employed to ensure dataset quality.

\subsection{Data Source Composition and Train/Test Split}

As described in Section~\ref{sec:3} of the main paper, TickTockVQA is collected from seven diverse sources. Table~\ref{tab:split_by_source} provides the complete breakdown of sample counts per source and their assignment to train/test splits. 

\subsection{Temporal Distribution Analysis}

\begin{figure}
    \centering
    \includegraphics[width=0.95\linewidth]{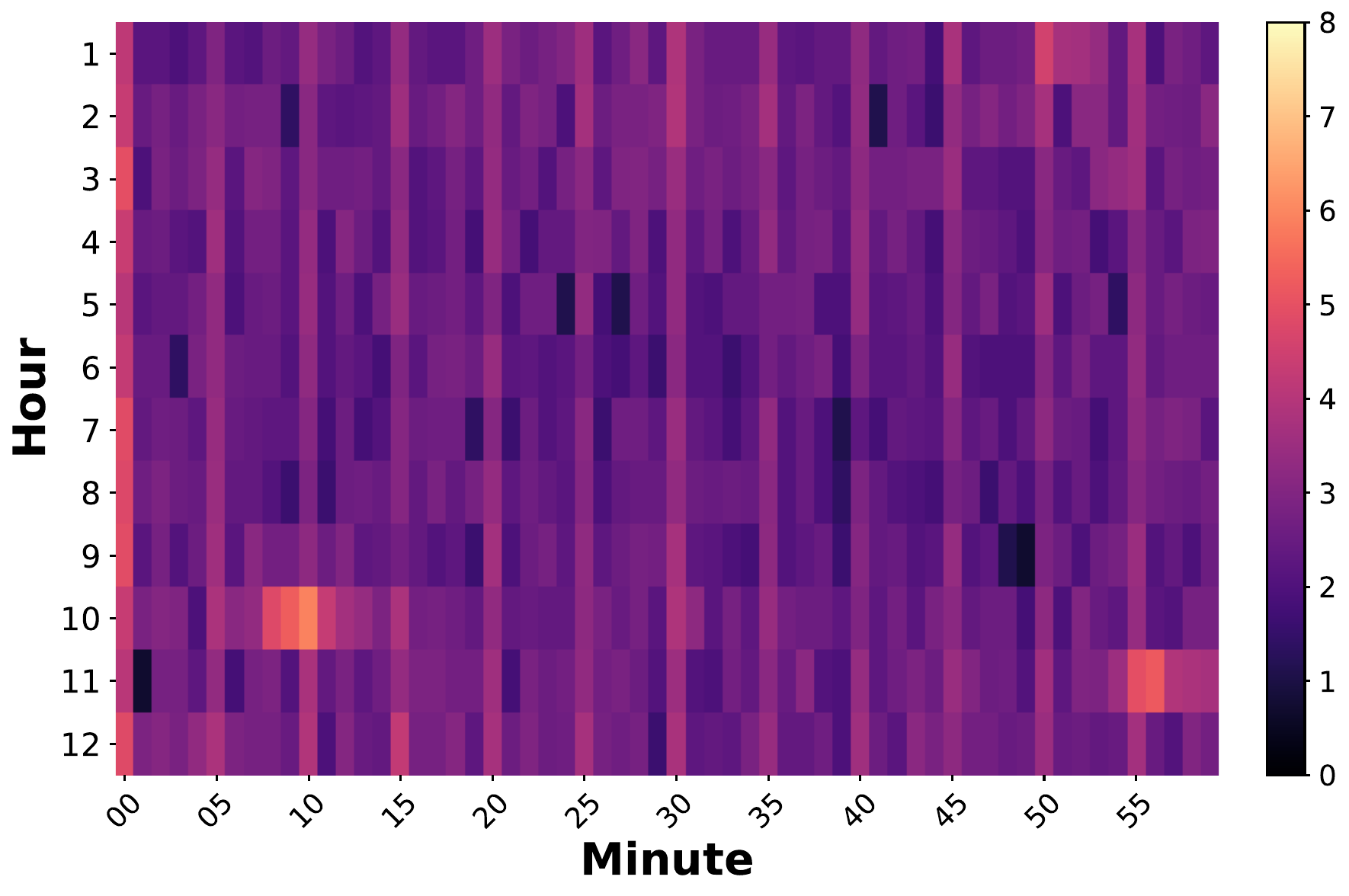}
    \caption{\textbf{Clock annotation density heatmap.} Distribution of labeled times across all hours (1--12) and minutes (0--59). Color intensity represents $\ln(1 + \text{count})$ to balance visibility across frequency ranges. Darker regions indicate higher sample density, with notable concentration around 10:10.}
    \label{fig:clock_heatmap}
\end{figure}

We analyze the distribution of annotated times across all 12 hours and 60 minutes to characterize both inherent biases and the coverage achieved through our filtering pipeline. Figure~\ref{fig:clock_heatmap} presents a two-dimensional heatmap showing the density of labeled times. The distribution is generally uniform, with noticeable concentration around aesthetically preferred times such as 10:10. This bias reflects the prevalence of such times in product photography and stock images.

\subsection{Marginal Distribution Analysis}

\begin{figure}
    \centering
    \begin{subfigure}{0.48\linewidth}
        \centering
        \includegraphics[width=\linewidth]{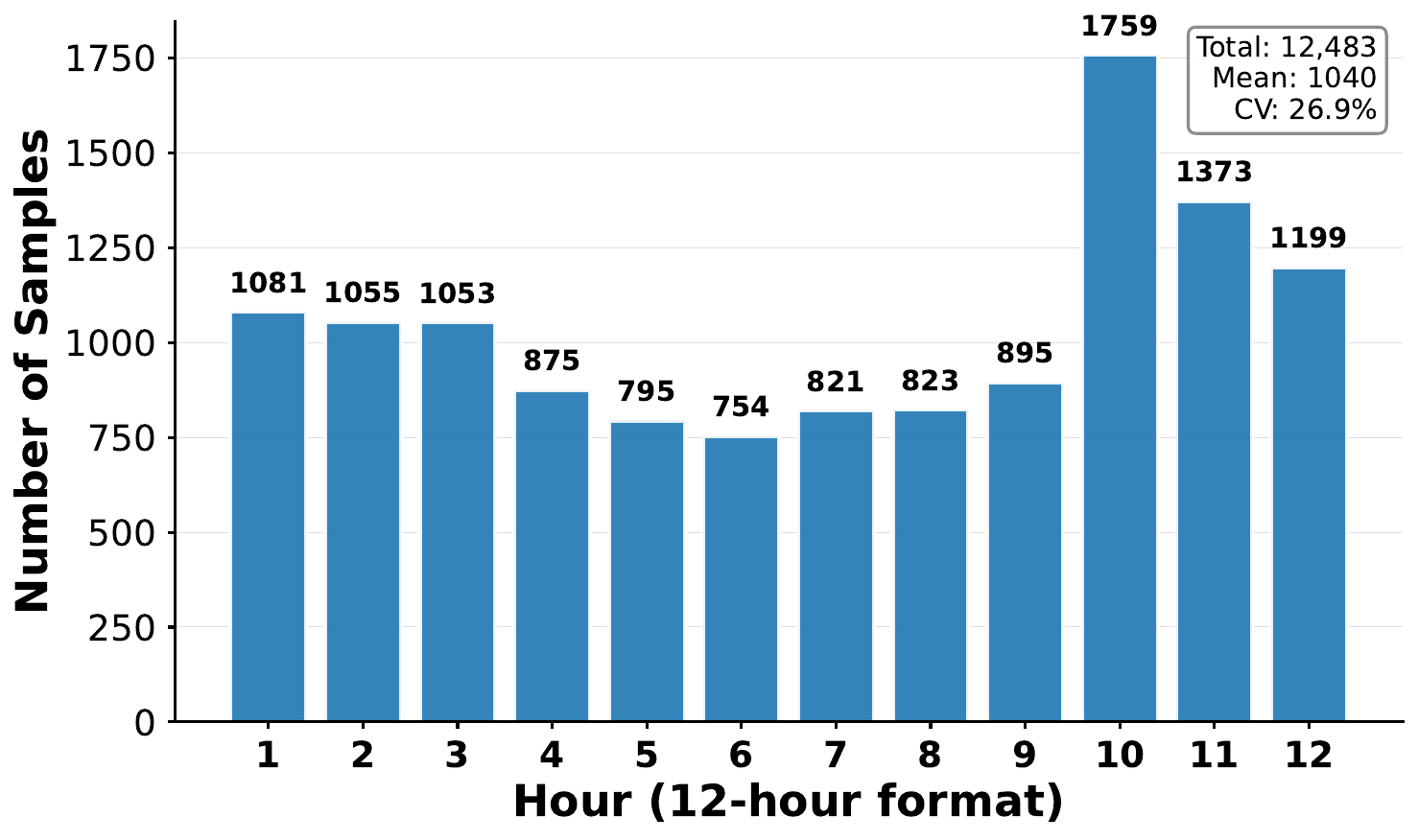}
        \caption{Hour distribution (1--12)}
        \label{fig:clock_hour_dist}
    \end{subfigure}
    \hfill
    \begin{subfigure}{0.48\linewidth}
        \centering
        \includegraphics[width=\linewidth]{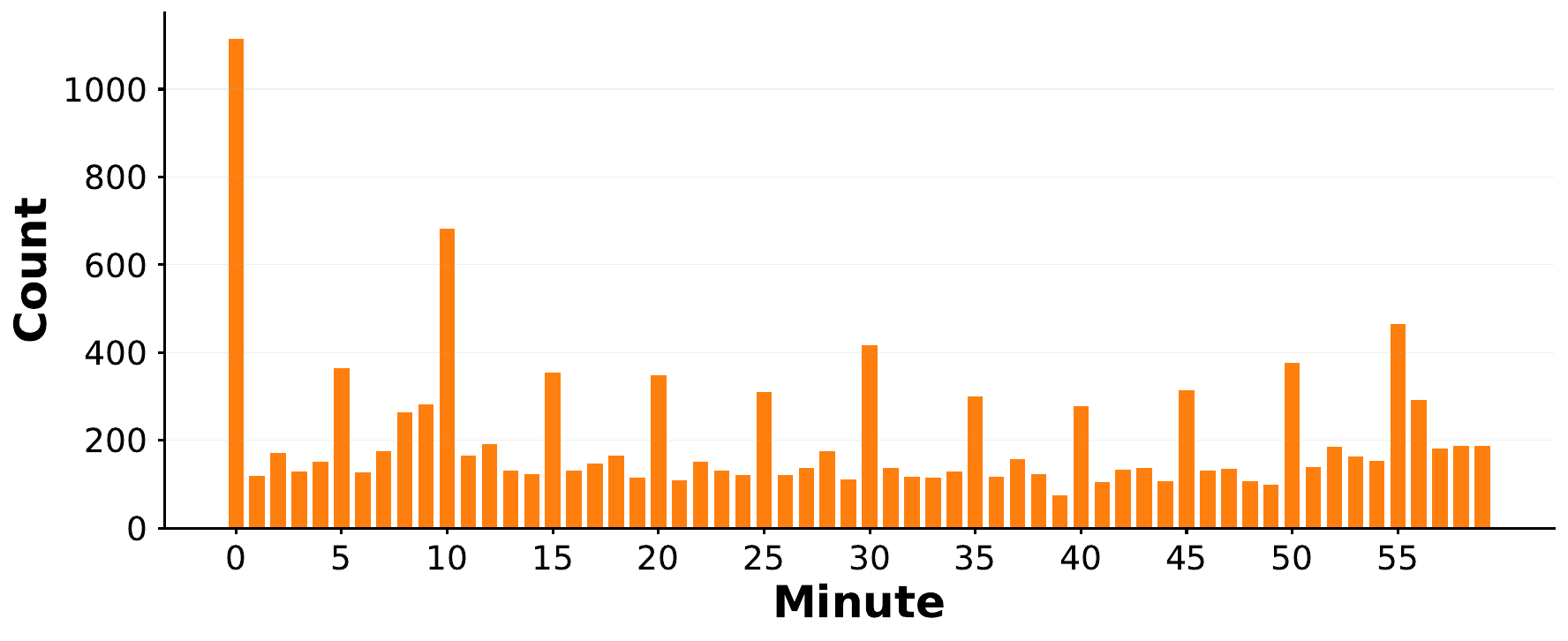}
        \caption{Minute distribution (0--59)}
        \label{fig:clock_minute_dist}
    \end{subfigure}
    \caption{\textbf{Marginal temporal distributions.} (a) Hour distribution shows reasonable balance with CV=26.9\%. Hour 10 is overrepresented due to the 10:10 aesthetic bias. (b) Minute distribution reveals expected peaks at canonical positions (0, 15, 30, 45) but maintains coverage across all 60 minutes.}
    \label{fig:temporal_marginals}
\end{figure}

Figure~\ref{fig:temporal_marginals} provides a detailed breakdown of hour and minute distributions. The hour distribution (Figure~\ref{fig:clock_hour_dist}) shows that hours 10, 11, and 12 are slightly overrepresented due to the 10:10 bias. However, all hours retain substantial coverage (minimum 754 samples for hour 6, maximum 1,759 for hour 10), with a coefficient of variation of 26.9\%, indicating reasonable balance.

The minute distribution (Figure~\ref{fig:clock_minute_dist}) reveals that canonical positions (0, 10, 15, 30, 45) occur more frequently than arbitrary minutes. Our filtering process substantially reduces these imbalances compared to raw web-crawled data, but residual skew toward common clock hand positions remains. Critically, all 60 minutes are represented in the dataset, ensuring coverage of fine-grained temporal reading challenges.

\begin{table}
    \centering
    \small
    \caption{\textbf{TickTockVQA data source composition and train/test split.} Only COCO, Open Images, and Clock Movies are used for testing; all other sources are reserved for training. This separation ensures evaluation on out-of-distribution sources.}
    \label{tab:split_by_source}
    \begin{tabular}{lcc}
        \toprule
        \textbf{Source} & \textbf{Images} & \textbf{Split} \\
        \midrule
        \multicolumn{3}{l}{\textit{Test Sources}} \\
        COCO                 & 2,063 & Test \\
        Open Images (OID)    & 1,940 & Test \\
        Clock Movies         & 1,244 & Test \\
        \midrule
        \multicolumn{3}{l}{\textit{Training Sources}} \\
        Visual Genome        & 1,246 & Train \\
        SBU Captions         & 1,533 & Train \\
        CC12M                & 3,677 & Train \\
        ImageNet             &   780 & Train \\
        \midrule
        \textbf{Total (Test)}  & \textbf{5,247} & \textbf{Test} \\
        \textbf{Total (Train)} & \textbf{7,236} & \textbf{Train} \\
        \textbf{Grand Total}   & \textbf{12,483} & \textbf{All} \\
        \bottomrule
    \end{tabular}
\end{table}

\section{Performance Analysis Across Clock Types and Conditions}
\label{sec:performance_analysis}

This section provides granular performance analysis of our ITGR model across different clock types, environmental conditions, and design variations. These analyses reveal which factors most significantly impact clock reading accuracy.

\subsection{Performance by Clock Type}

\begin{figure}
    \centering
    \includegraphics[width=0.95\linewidth]{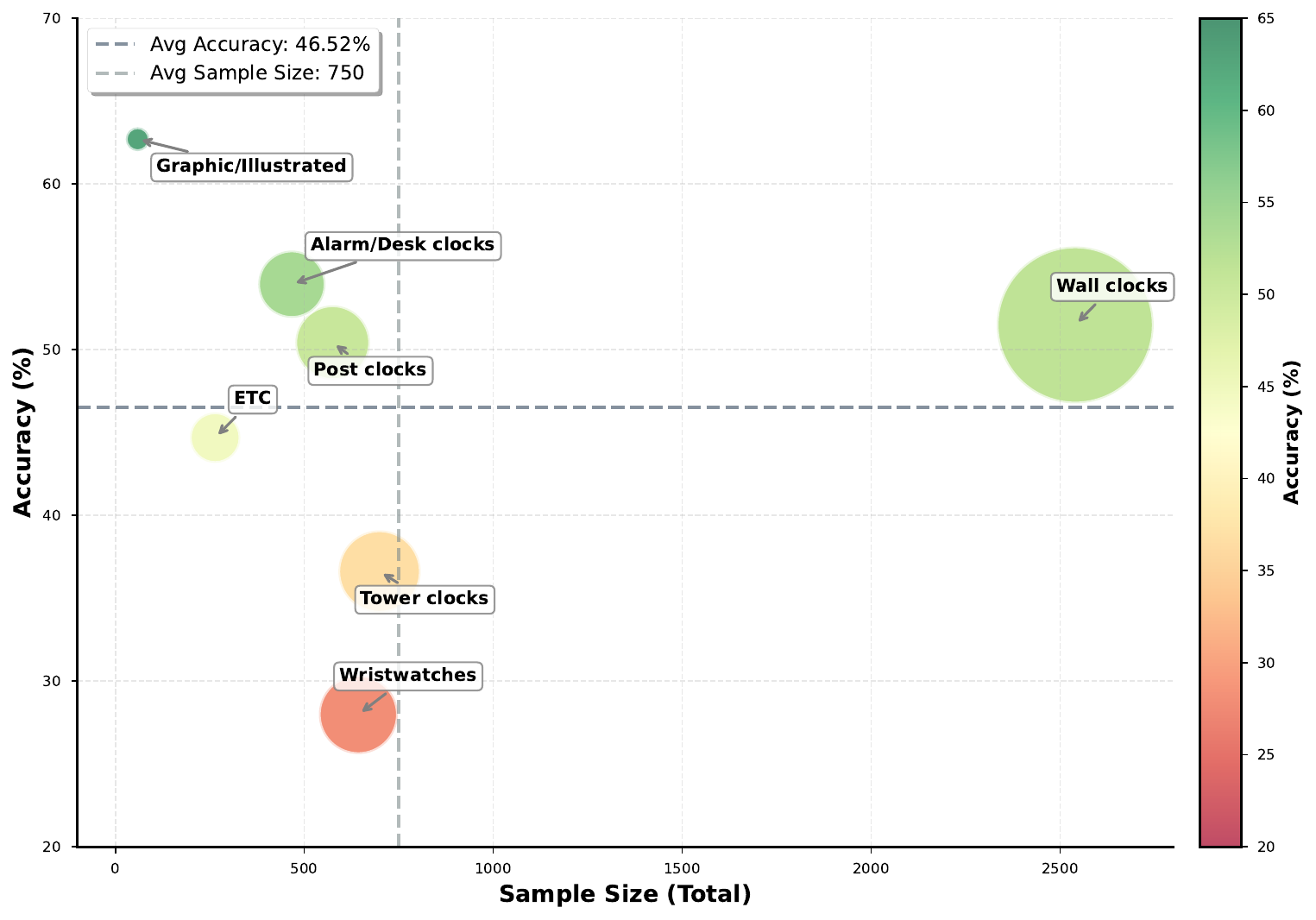}
    \caption{\textbf{ITGR accuracy breakdown by clock type.} Performance varies significantly across categories. Graphic/Illustrated clocks achieve the highest accuracy (62.71\%) due to high contrast and clean contours. Wristwatches show the lowest performance (27.99\%) due to small size, glass reflections, and occlusion. Bubble size represents sample count for each category. The dashed line indicates overall average accuracy (46.52\%).}
    \label{fig:itgr_clock_type}
\end{figure}

Figure~\ref{fig:itgr_clock_type} presents the breakdown of ITGR (Llama-3.2-11B with Swap-DPO) performance across seven clock categories. Performance varies dramatically, ranging from 27.99\% (wristwatches) to 62.71\% (graphic/illustrated clocks), revealing significant differences in task difficulty.

\textbf{Key Observations:}

\begin{itemize}[leftmargin=*,noitemsep]
    \item \textbf{Graphic/Illustrated clocks (62.71\%):} Highest performance due to high contrast, clean contours, and minimal background clutter. These clocks typically appear in controlled settings with frontal viewpoints.
    
    \item \textbf{Wristwatches (27.99\%):} Lowest performance despite substantial training data (1,238 samples). Challenges include: (1) small clock face size in images, (2) glass reflections obscuring hands, (3) depth-of-field blur, (4) curved surfaces causing distortion, and (5) frequent hand overlap at small scales.
    
    \item \textbf{Wall clocks (50.60\%):} Despite being the largest category (4,046 samples), performance is moderate. This indicates that simply scaling data does not guarantee improved performance; visual complexity in real-world wall clock scenarios (varied lighting, viewing angles, occlusion) poses persistent challenges.
    
    \item \textbf{Tower clocks (44.66\%):} Moderate performance. Challenges include extreme viewing angles, atmospheric effects, and distance-related image quality degradation.
    
    \item \textbf{Alarm/Desk clocks (47.63\%):} Performance similar to wall clocks, benefiting from typically frontal viewing angles but challenged by reflective surfaces and small digital displays that can distract the model.
\end{itemize}

\textbf{Implications:} The 35pp performance gap between the easiest and hardest categories demonstrates that clock reading difficulty depends heavily on physical form factor and imaging conditions, not merely on the number of training samples.

\subsection{Performance by Environmental Conditions, Transformations, and Design}

\begin{figure*}[t]
    \centering
    \includegraphics[width=0.95\linewidth]{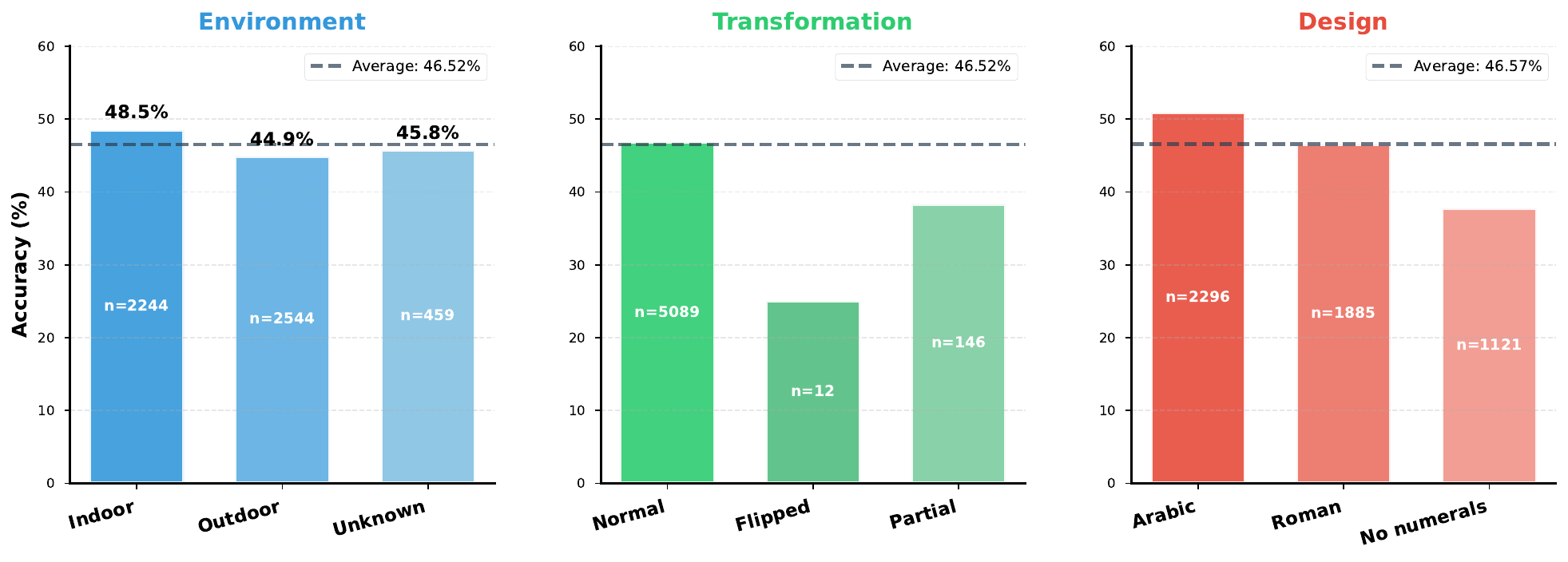}
    \caption{\textbf{ITGR accuracy breakdown across three categorical dimensions.} (a) \textbf{Environment:} Performance is stable across indoor (48.5\%), outdoor (44.9\%), and unknown (45.8\%) settings, indicating robustness to background context and lighting variation. (b) \textbf{Transformation:} Severe degradation for flipped clocks (23.1\%), revealing fragility to unusual orientations. Partial occlusion (37.3\%) also degrades performance. (c) \textbf{Design:} Performance is consistent across Arabic (48.2\%) and Roman (46.8\%) numerals but degrades for clocks without numerals (36.2\%), suggesting reliance on numerical markers for spatial reference. Dashed lines indicate overall average accuracy.}
    \label{fig:itgr_category_analysis}
\end{figure*}

Figure~\ref{fig:itgr_category_analysis} decomposes ITGR performance across three categorical dimensions: (a) environment (indoor/outdoor/unknown), (b) geometric transformation (normal/flipped/partial), and (c) clock face design (Arabic/Roman/no numerals).

\textbf{Environmental Robustness (Figure~\ref{fig:itgr_category_analysis}a):}
The model demonstrates stable performance across different environmental settings: indoor (48.5\%, n=2,244), outdoor (44.9\%, n=2,544), and unknown (45.8\%, n=459). This 3.6pp variation suggests that background context, ambient lighting, and scene clutter alone do not significantly destabilize predictions. The model has learned to focus on the clock itself rather than being distracted by environmental factors.

\textbf{Transformation Sensitivity (Figure~\ref{fig:itgr_category_analysis}b):}
The model exhibits severe degradation for transformed clocks:
\begin{itemize}[leftmargin=*,noitemsep]
    \item \textbf{Normal orientation (46.9\%, n=5,089):} Baseline performance
    \item \textbf{Flipped/rotated (23.1\%, n=12):} 50\% relative performance drop, indicating brittleness to non-canonical orientations. The model struggles when clocks are horizontally flipped or rotated, suggesting it has learned orientation-specific features rather than rotation-invariant representations.
    \item \textbf{Partial/occluded (37.3\%, n=146):} 20\% relative drop, showing sensitivity to missing information even when hands remain visible.
\end{itemize}

\textbf{Design Dependency (Figure~\ref{fig:itgr_category_analysis}c):}
Performance varies by clock face design:
\begin{itemize}[leftmargin=*,noitemsep]
    \item \textbf{Arabic numerals (48.2\%, n=2,296):} Slightly better, likely due to clearer spatial references
    \item \textbf{Roman numerals (46.8\%, n=1,885):} Comparable performance, indicating successful generalization across numeral systems
    \item \textbf{No numerals (36.2\%, n=1,121):} 25\% relative drop, revealing reliance on hour markers for spatial reasoning. Without explicit markers, the model must infer positions purely from hand angles, which is more challenging.
\end{itemize}

\subsection{Key Findings and Implications}

\begin{enumerate}[leftmargin=*,noitemsep]
    \item \textbf{Environmental robustness vs. structural fragility:} While ITGR is robust to environmental variation (lighting, background), it remains highly sensitive to structural deviations (flipped orientation, missing numerals). This suggests that future work should focus on geometric augmentation and rotation-equivariant architectures.
    
    \item \textbf{Data scaling is insufficient:} Wall clocks, despite having the most training samples (4,046), achieve only moderate accuracy (50.60\%). This confirms that visual complexity and physical form factor are more important than sample quantity.
    
    \item \textbf{Form factor matters most:} The 35pp gap between clock types (wristwatches vs. graphic clocks) far exceeds the 4pp gap from environmental conditions, indicating that physical form factor is the dominant difficulty factor.
    
    \item \textbf{Future directions:} Improving performance on low-accuracy categories (wristwatches, flipped clocks, no-numeral designs) through specialized data augmentation, multi-scale processing, and rotation-invariant features represents a promising research direction.
\end{enumerate}


\section{Effect of Swap-DPO on Hand Confusion}
\label{sec:swap_dpo_analysis}

This section analyzes how our Swap-DPO method specifically addresses the hand-swapping problem through targeted preference learning. We compare three training strategies: (1) SFT only, (2) Random-DPO (baseline DPO with random error correction), and (3) Swap-DPO (our proposed method).

\subsection{Experimental Setup}

For each model (Qwen2.5-VL-7B, Gemma3-12B, Llama-3.2-11B), we train three variants:
\begin{itemize}[leftmargin=*,noitemsep]
    \item \textbf{SFT:} Supervised fine-tuning on TickTockVQA training set (7,236 samples) for 10 epochs
    \item \textbf{Random-DPO:} SFT + DPO with randomly selected incorrect predictions as rejected responses
    \item \textbf{Swap-DPO:} SFT + DPO with geometrically swapped times as rejected responses (our method)
\end{itemize}

We evaluate each variant under two metrics:
\begin{itemize}[leftmargin=*,noitemsep]
    \item \textbf{Baseline (B):} Standard full-time accuracy (hour and minute must both be correct)
    \item \textbf{Swap-equivalence (S):} Accuracy when allowing hour/minute hand swaps (e.g., predicting 06:18 for ground truth 03:30 counts as correct).
\end{itemize}

The gap $\Delta = S - B$ directly quantifies hand-swap confusion: a larger gap indicates more frequent role reversal errors.

\subsection{Quantitative Results}

Table~\ref{tab:stagewise_results} presents full-time accuracy across all three training strategies. The results consistently demonstrate that Swap-DPO reduces hand-swap confusion while improving overall accuracy.

\begin{table*}[h]
    \centering
    \small
    \caption{\textbf{Full-time accuracy across SFT, Random-DPO, and Swap-DPO.} B denotes Baseline accuracy (strict), S denotes Swap-equivalence accuracy (allowing hand swaps), and $\Delta$ is the hand-swap gap ($S - B$). A smaller $\Delta$ indicates less hand confusion. Swap-DPO consistently reduces $\Delta$ compared to both SFT and Random-DPO while improving overall accuracy.}
    \label{tab:stagewise_results}
    \begin{tabular}{lccccccccc}
        \toprule
        & \multicolumn{3}{c}{\textbf{SFT}} 
        & \multicolumn{3}{c}{\textbf{Random-DPO}}
        & \multicolumn{3}{c}{\textbf{Swap-DPO}} \\
        \cmidrule(lr){2-4} \cmidrule(lr){5-7} \cmidrule(lr){8-10}
        \textbf{Model} 
        & \textbf{B} & \textbf{S} & \boldmath$\Delta$  
        & \textbf{B} & \textbf{S} & \boldmath$\Delta$
        & \textbf{B} & \textbf{S} & \boldmath$\Delta$ \\
        \midrule
        Qwen2.5-VL-7B   & 20.3 & 22.8 & +2.42
               & 20.6 & 23.4 & +2.75
               & \textbf{23.1} & \textbf{25.1} & \textbf{+2.02} \\
        Gemma3-12B  & 34.2 & 37.1 & +2.90
               & 35.0 & 38.0 & +3.00
               & \textbf{35.3} & \textbf{37.9} & \textbf{+2.57} \\
        Llama-3.2-11B  & 45.8 & 48.1 & +2.32
               & 45.6 & 47.7 & +2.06
               & \textbf{46.2} & \textbf{48.5} & \textbf{+2.26} \\
        \midrule
        \multicolumn{10}{l}{\textit{Average across models:}} \\
        Mean & 33.4 & 36.0 & +2.55 & 33.7 & 36.4 & +2.60 & \textbf{34.9} & \textbf{37.2} & \textbf{+2.28} \\
        \bottomrule
    \end{tabular}
\end{table*}

\subsection{Key Observations}

\textbf{1. SFT establishes strong baseline but exhibits hand confusion:}
Supervised fine-tuning substantially improves performance across all models (e.g., Llama: 1.41\% zero-shot → 45.8\% SFT). However, a persistent 2.32--2.90pp gap between B and S metrics indicates that 5--7\% of errors are pure hand-swap mistakes where the model has correctly localized both hands but assigned incorrect semantic roles.

\textbf{2. Random-DPO fails to reduce hand confusion:}
Surprisingly, applying standard DPO with randomly selected incorrect predictions as rejected responses \textit{increases} the hand-swap gap in 2 out of 3 models (Qwen: +2.42 → +2.75, Gemma: +2.90 → +3.00). This counterintuitive result suggests that generic error correction makes the model \textit{more} sensitive to hand ambiguity. We hypothesize that random negative samples lack the geometric consistency needed to teach hand role distinction; the model learns to avoid diverse errors but does not specifically learn which hand is which.

\textbf{3. Swap-DPO consistently reduces hand confusion:}
Our Swap-DPO method, which uses geometrically swapped times as rejected responses, achieves three critical improvements:
\begin{itemize}[leftmargin=*,noitemsep]
    \item \textbf{Reduced hand-swap gap:} Average $\Delta$ decreases from 2.55 (SFT) to 2.28 (Swap-DPO), a 10.6\% relative reduction. For Qwen, the reduction is 16.5\% (2.42 → 2.02).
    \item \textbf{Improved overall accuracy:} Baseline accuracy improves across all models (average: 33.4\% → 34.9\%), demonstrating that resolving hand confusion generalizes to other error types.
    \item \textbf{Consistency across architectures:} Swap-DPO outperforms Random-DPO on all three models, indicating robustness of the approach.
\end{itemize}

\subsection{Why Does Swap-DPO Work?}

The effectiveness of Swap-DPO stems from its geometric consistency:
\begin{enumerate}[leftmargin=*,noitemsep]
    \item \textbf{Contrastive hand role learning:} By presenting the model with two geometrically plausible interpretations of the same clock (correct vs. swapped), we force it to learn which visual features (hand length, thickness, position) correspond to which semantic role (hour vs. minute).
    
    \item \textbf{Hard negative mining:} Swapped times are "hard negatives" because they are geometrically consistent with the visual input but semantically incorrect. This is more informative than random wrong times, which may be geometrically implausible.
    
    \item \textbf{Explicit disambiguation signal:} Unlike SFT, which only provides positive examples, Swap-DPO explicitly teaches what \textit{not} to predict, specifically targeting the most common failure mode.
\end{enumerate}

\subsection{Limitations and Remaining Challenges}

Despite Swap-DPO's improvements, a 2.0--2.6pp hand-swap gap persists, indicating that 4--6\% of errors remain pure hand-swap confusions. This suggests:
\begin{itemize}[leftmargin=*,noitemsep]
    \item \textbf{Ambiguous cases:} Some clocks have nearly identical hand lengths or poor image quality, making disambiguation genuinely difficult even for humans.
    \item \textbf{Model capacity:} Current VLM architectures may lack sufficient fine-grained spatial reasoning capabilities to perfectly distinguish hands in all scenarios.
    \item \textbf{Dataset bias:} The 2--3\% residual gap may represent an upper bound given inherent ambiguities in real-world analog clocks.
\end{itemize}

Future work could explore: (1) multi-stage reasoning (explicit hand detection → role assignment → time reading), (2) uncertainty quantification to flag ambiguous cases, and (3) contrastive pre-training on synthetic clock data with perfect hand labels.


\section{Implementation Details}
\label{sec:implementation}

This section provides comprehensive implementation details to ensure full reproducibility of our experiments. We report all hyperparameters, training configurations, and computational requirements for the three VLM backbones used in our study.

\subsection{DPO Training Configuration}
\label{sec:dpo_config}

Table~\ref{tab:dpo_hyperparameters} summarizes the complete DPO training configuration across all three model architectures. We employ a consistent training strategy with minor architecture-specific adjustments to accommodate different model characteristics.

\begin{table*}[p]
\centering
\caption{\textbf{Complete DPO training hyperparameters and configurations.} We report all settings used for Direct Preference Optimization across three VLM backbones. Model-specific differences are highlighted. All models use 8$\times$ NVIDIA A6000 GPUs (48GB each).}
\label{tab:dpo_hyperparameters}
\resizebox{\textwidth}{!}{
\begin{tabular}{l@{\hspace{6pt}}c@{\hspace{6pt}}c@{\hspace{6pt}}c@{\hspace{6pt}}p{0.25\textwidth}}
\toprule
\textbf{Configuration} & \textbf{Qwen2.5-VL-7B} & \textbf{Llama-3.2-11B} & \textbf{Gemma3-12B} & \textbf{Notes} \\
\midrule
\multicolumn{5}{l}{\textit{\textbf{DPO-Specific Parameters}}} \\
\midrule
Loss function & sigmoid & sigmoid & sigmoid & Standard DPO loss \\
$\beta$ (temperature) & 0.3 & 0.3 & 0.3 & Controls preference strength \\
Precompute ref. logprobs & false & false & false & Compute on-the-fly \\
\midrule
\multicolumn{5}{l}{\textit{\textbf{LoRA Configuration}}} \\
\midrule
LoRA enabled & \xmark & \cmark & \cmark & Qwen uses full fine-tuning \\
LoRA rank ($r$) & --- & 64 & 64 & --- \\
LoRA alpha ($\alpha$) & --- & 64 & 64 & $\alpha = r$ for stability \\
LoRA dropout & --- & 0.05 & 0.05 & --- \\
Target modules & --- & all linear & all linear & Except embeddings/LM head \\
Vision LoRA & --- & \cmark & \cmark & Apply LoRA to vision tower \\
DoRA & --- & \xmark & \xmark & Standard LoRA \\
\midrule
\multicolumn{5}{l}{\textit{\textbf{Batch Size \& Parallelization}}} \\
\midrule
Global batch size & 256 & 256 & 256 & Effective batch size \\
Batch per device & 4 & 8 & 4 & Per-GPU batch size \\
Gradient accum. steps & 8 & 4 & 8 & $= 256 / (\text{batch} \times \text{GPUs})$ \\
Num. devices & 8 & 8 & 8 & NVIDIA A6000 (48GB) \\
\midrule
\multicolumn{5}{l}{\textit{\textbf{Optimization Hyperparameters}}} \\
\midrule
Num. epochs & 4 & 4 & 4 & Consistent across models \\
Learning rate (LLM) & 2e-6 & 2e-6 & 2e-6 & Base LLM learning rate \\
Learning rate (vision) & 2e-6 & 2e-6 & 2e-6 & Vision tower learning rate \\
Learning rate (projector) & 1e-5 & 1e-5 & 1e-5 & 5$\times$ higher for projector \\
Weight decay & 0.1 & 0.1 & 0.1 & AdamW regularization \\
Adam $\beta_1$ & 0.9 & 0.9 & 0.9 & Default \\
Adam $\beta_2$ & 0.95 & 0.95 & 0.95 & Slightly lower than default \\
Warmup ratio & 0.03 & 0.03 & 0.03 & 3\% of total steps \\
LR scheduler & cosine & cosine & cosine & Cosine annealing to 0 \\
\midrule
\multicolumn{5}{l}{\textit{\textbf{Memory \& Precision}}} \\
\midrule
Mixed precision & bfloat16 & bfloat16 & bfloat16 & Training dtype \\
FP16 & \xmark & \xmark & \xmark & Use bfloat16 instead \\
TF32 & \cmark & \cmark & \cmark & NVIDIA Ampere+ acceleration \\
Gradient checkpointing & \cmark & \cmark & \cmark & Recompute activations \\
DeepSpeed stage & ZeRO-3 & ZeRO-3 & ZeRO-3 & Partition optimizer states \\
Flash Attention 2 & \cmark & \cmark & \xmark & Gemma3: eager attention \\
Liger kernel & \cmark & \cmark & \cmark & Fused RMSNorm + cross-entropy \\
\midrule
\multicolumn{5}{l}{\textit{\textbf{Module Freezing Strategy}}} \\
\midrule
Freeze vision tower & \xmark & \cmark & \cmark & Qwen: full fine-tuning \\
Freeze LLM & \xmark & \cmark & \cmark & Qwen: full fine-tuning \\
Freeze projector & \xmark & \xmark & \xmark & Always trainable \\
\bottomrule
\end{tabular}
}
\end{table*}

\textbf{Architecture-Specific Configuration Notes:}

\textbf{(1) Qwen2.5-VL-7B:} We apply full fine-tuning (no LoRA) due to its relatively small size (7B parameters) and efficient architecture. Qwen uses dynamic resolution processing with configurable min/max pixels (401K--1003K), allowing adaptive handling of various image sizes. Flash Attention 2 is enabled for memory efficiency. The model's native support for variable-resolution inputs eliminates the need for fixed-size preprocessing.

\textbf{(2) Llama-3.2-11B:} We employ LoRA (rank 64, alpha 64) on all linear layers including the vision tower to reduce memory footprint. The larger per-device batch size (8 vs. 4 for Qwen/Gemma) is possible due to LoRA's parameter efficiency—only $\sim$2\% of parameters are trainable. Lazy preprocessing (on-the-fly image loading) accelerates training. Flash Attention 2 is supported and enabled.

\textbf{(3) Gemma3-12B:} Similar to Llama, we use LoRA (rank 64, alpha 64) for memory efficiency. However, we use \textit{eager attention} instead of Flash Attention 2, as recommended by the Gemma3 technical report due to numerical stability concerns with certain attention patterns. DoRA (weight-decomposed LoRA) is disabled for training stability. We disable lazy preprocessing to ensure deterministic image loading order.

\textbf{Common Configuration Rationale:} All models share core settings: sigmoid DPO loss with $\beta=0.3$ (stronger preference signal than default 0.1), global batch size of 256 (necessary for stable DPO training), and 4 training epochs (sufficient for convergence without overfitting). We use ZeRO-3 (partitioned optimizer states and gradients) with bfloat16 mixed precision and gradient checkpointing to enable training on 8$\times$ A6000 GPUs. The projector module always receives a 5$\times$ higher learning rate (1e-5 vs. 2e-6) because it bridges frozen/slowly-adapted vision features to the language model and requires faster adaptation.

\subsection{DPO Preference Data Generation}
\label{sec:dpo_data_generation}

We automatically generate DPO preference pairs by running inference with the SFT model on the training set. This section provides implementation details and configuration parameters beyond what is described in the main paper (Algorithm~\ref{alg:main_summary}).

\begin{table}[h]
\centering
\caption{\textbf{DPO preference data generation configuration.}}
\label{tab:dpo_data_config}
\begin{tabular}{lc}
\toprule
\textbf{Parameter} & \textbf{Value} \\
\midrule
\multicolumn{2}{l}{\textit{SFT Checkpoint Selection}} \\
Qwen2.5-VL-7B & checkpoint-145 (epoch 5) \\
Llama-3.2-11B & checkpoint-174 (epoch 6) \\
Gemma3-12B & checkpoint-203 (epoch 7) \\
\midrule
\multicolumn{2}{l}{\textit{SFT Accuracy at Selected Checkpoint}} \\
Qwen2.5-VL-7B & 20.3\% Full Time Acc \\
Llama-3.2-11B & 45.8\% Full Time Acc \\
Gemma3-12B & 34.2\% Full Time Acc \\
\midrule
Training samples & 7,236 \\
Minute tolerance & $\pm$2 minutes \\
\bottomrule
\end{tabular}
\end{table}

\subsubsection{Inference Configuration for Data Generation}

We run SFT model inference on all 7,236 training images using the following configuration:

\begin{itemize}[leftmargin=*,noitemsep]
    \item \textbf{Batch size:} 16 (same as training)
    \item \textbf{Temperature:} 0.0 (greedy decoding for deterministic outputs)
    \item \textbf{Max new tokens:} 16 (sufficient for ``HH:MM'' format)
    \item \textbf{Prompt:} Inference prompt (Table~\ref{tab:inference_prompt})
    \item \textbf{Hardware:} 8$\times$ A6000 GPUs with DeepSpeed inference
    \item \textbf{Time:} $\sim$2 hours per model
\end{itemize}

\subsubsection{Swap-DPO Transformation: Edge Cases}

The main paper describes the SwapHands transformation. Here we document edge cases and implementation details:

\textbf{1. Times near 12:00:}
\begin{itemize}[leftmargin=*,noitemsep]
    \item Input: 12:00 $\rightarrow$ Output: 12:00 (degenerate case, both hands point up)
    \item \textit{Handling:} These cases yield degenerate swaps (i.e., $\textsc{SwapHands}(y_{\text{gt}})=y_{\text{gt}}$) and are filtered by our validation checks (Sec.~~\ref{sec:qc-validation}) when swap-based negatives are used.
\end{itemize}

\textbf{2. Near-overlapping hands:}
\begin{itemize}
\item Example: $1{:}05$ ($\theta_h=32.5^\circ$, $\theta_m=30^\circ$)
\item Swapped: $1{:}05$ (nearly identical)
\item Handling: Such cases are filtered out by our distinctness / temporal-distance checks Section.~\ref{sec:qc-validation}
\end{itemize}

\textbf{3. Half-hour positions:}
\begin{itemize}[leftmargin=*,noitemsep]
    \item Example: 3:30 $\rightarrow$ Swapped: 6:18
    \item Hour hand at 105° (halfway between 3 and 4)
    \item Minute hand at 180° (pointing at 6)
    \item \textit{Handling:} Works as intended
\end{itemize}

\textbf{4. Rounding behavior:}
\begin{itemize}[leftmargin=*,noitemsep]
    \item We use floor division for hours: $h_{\text{new}} = \lfloor \theta_m / 30 \rfloor$
    \item We use modulo for minutes: $m_{\text{new}} = (\theta_h / 6) \bmod 60$
    \item This ensures outputs remain in valid ranges: $h \in [0,11]$, $m \in [0,59]$
\end{itemize}

\subsubsection{Quality Control and Validation}\label{sec:qc-validation}

We perform the following validation checks on generated preference pairs:

\begin{enumerate}[leftmargin=*,noitemsep]
    \item \textbf{Format validation:} Both $y_w$ and $y_l$ must be valid HH:MM strings
    \item \textbf{Distinctness:} $y_w \neq y_l$ (reject if swapped time equals ground truth)
    \item \textbf{Geometric plausibility:} For Swap-DPO pairs, verify that swapped time corresponds to a valid clock configuration
    \item \textbf{Temporal distance:} Ensure $|y_w - y_l| > 5$ minutes to avoid noisy signals from nearly identical times
\end{enumerate}

After validation, we retain 7,187 out of 7,236 samples (99.3\%). The 49 rejected samples include parse failures, degenerate swaps, and format errors.

\subsubsection{Data Storage and Format}

The final preference dataset is stored as a JSON Lines file where each line contains:

\begin{itemize}[leftmargin=*,noitemsep]
    \item \texttt{image\_path}: Relative path to training image
    \item \texttt{chosen}: Ground truth time (always $y_w$)
    \item \texttt{rejected}: Constructed negative sample ($y_l$)
\end{itemize}

\subsubsection{Comparison with Random-DPO Baseline}

To validate our hybrid strategy, we compare against a \textit{Random-DPO} baseline where $y_l$ is sampled uniformly from incorrect times (excluding ground truth). As shown in Table~\ref{tab:stagewise_results}, Random-DPO yields a hand-swap gap of +2.60\% (averaged across models), compared to +2.28\% for our hybrid approach.

\textbf{Key Finding:} The hybrid strategy achieves the best hand-swap gap reduction (+2.28\%) by combining:
\begin{itemize}[leftmargin=*,noitemsep]
    \item \textbf{Geometric consistency} from Swap-DPO (teaches hand roles)
    \item \textbf{Error diversity} from SFT mistakes (teaches robustness)
\end{itemize}

Pure Swap-DPO (using swapped times for all samples, even when SFT is wrong) performs slightly worse (+2.15\% gap) because it ignores the model's natural error distribution, missing opportunities to correct systematic mistakes like occlusion handling or numeral misreading. Random-DPO performs worst (+2.60\% gap) because randomly sampled times lack geometric consistency and fail to specifically target hand confusion.


\section{Prompt Engineering and Design}
\label{sec:prompt_design}

Effective prompt design is critical for teaching VLMs to read analog clocks accurately. This section describes our comprehensive prompting strategy, including training-time prompt rotation, inference-time simplification.

\subsection{Training Prompt Design and Rotation Strategy}

During supervised fine-tuning, we employ \textit{prompt rotation}—cycling through three semantically equivalent but lexically diverse prompts to prevent overfitting to specific phrasings. Table~\ref{tab:training_prompts} presents our training prompts with key phrase variations highlighted.

\begin{table*}[h]
\centering
\caption{\textbf{Training prompt variations for supervised fine-tuning.} We rotate through three semantically equivalent prompts to prevent overfitting while maintaining consistent instruction semantics. Key phrase variations are shown; all prompts share the core instructions for hand identification, ambiguity resolution, and output formatting.}
\label{tab:training_prompts}
\small
\begin{tabular}{lp{0.78\textwidth}}
\toprule
\textbf{Prompt ID} & \textbf{Full Prompt Text} \\
\midrule
\textbf{Prompt A} & 
\texttt{\textcolor{blue}{Identify} the \textcolor{blue}{single most prominent} analog clock in the image (ignore digital displays). If multiple clocks are visible, choose the largest clearly visible face. Read hour = \textcolor{red}{shorter thicker hand} and minute = \textcolor{red}{longer thinner hand}; ignore any seconds hand. If a hand lies between marks, use the lower hour and the nearest minute. Output only HH:MM (12-hour, leading zero, no AM/PM). If no analog clock is visible, output NO CLOCK.} \\
\midrule
\textbf{Prompt B} & 
\texttt{\textcolor{blue}{Find} the \textcolor{blue}{primary} analog clock (exclude digital). Select the largest clearly visible face when multiple are present. Hour = \textcolor{red}{shorter thicker hand}; minute = \textcolor{red}{longer thinner hand}; ignore seconds. If a hand is between ticks, choose the lower hour and nearest minute. Answer strictly as HH:MM (12-hour, leading zero). If no analog clock is found, answer NO CLOCK.} \\
\midrule
\textbf{Prompt C} & 
\texttt{\textcolor{blue}{Locate} the \textcolor{blue}{most visible} analog clock and ignore any digital displays. If several clocks exist, pick the biggest clear dial. Use the \textcolor{red}{short thick hand} for hours and the \textcolor{red}{long thin hand} for minutes; ignore seconds. Between ticks: take the lower hour and nearest minute. Return only HH:MM (12-hour, leading zero). If none is present, return NO CLOCK.} \\
\midrule
\multicolumn{2}{l}{\textit{Common Core Instructions (present in all prompts):}} \\
\multicolumn{2}{l}{1. Select the most prominent analog clock (ignore digital displays)} \\
\multicolumn{2}{l}{2. Hour hand = short/thick; Minute hand = long/thin (ignore seconds)} \\
\multicolumn{2}{l}{3. Ambiguity resolution: if between marks, use lower hour and nearest minute} \\
\multicolumn{2}{l}{4. Output format: HH:MM (12-hour, leading zero, no AM/PM)} \\
\multicolumn{2}{l}{5. Fallback: output "NO CLOCK" if no analog clock visible} \\
\bottomrule
\end{tabular}
\end{table*}

\textbf{Prompt Design Rationale:}

\begin{enumerate}[leftmargin=*,noitemsep,topsep=3pt]
    \item \textbf{Explicit hand disambiguation:} All prompts explicitly describe hour hand attributes (\textit{short, thick}) and minute hand attributes (\textit{long, thin}). This addresses the core hand confusion problem by providing unambiguous semantic roles.
    
    \item \textbf{Multi-clock handling:} Instructions to select "the most prominent," "primary," or "most visible" clock ensure consistent behavior when multiple clocks appear in a single image ($\sim$8\% of training data). Without this, the model randomly attends to different clocks, causing training instability.
    
    \item \textbf{Ambiguity resolution rule:} The directive "if a hand is between marks, use the lower hour and nearest minute" provides a deterministic tie-breaking strategy. This reduces annotation ambiguity (annotators might disagree on 3:14 vs. 3:15) and training noise. This rule is consistent with standard clock reading: if the hour hand lies between two hour marks, we report the lower hour.
    
    \item \textbf{Structured output format:} Requiring "HH:MM" with 12-hour convention and leading zeros (e.g., "08:05" not "8:5") simplifies parsing and eliminates format-related errors. We use 12-hour format because most analog clocks display 1--12, not 0--23.
    
    \item \textbf{Fallback handling:} The "NO CLOCK" instruction prevents hallucination on images without visible analog clocks. This is critical for robustness on diverse web-scraped data where some images may be mislabeled or contain only digital clocks.
\end{enumerate}

\textbf{Rotation Strategy:} During training, we randomly sample one of the three prompts for each example with uniform probability (33.3\% each). This serves three purposes:

\begin{itemize}[leftmargin=*,noitemsep]
    \item \textbf{Prevents prompt memorization:} Forces the model to learn underlying task semantics rather than surface lexical patterns
    \item \textbf{Improves robustness:} Generalizes to unseen prompt formulations at test time
    \item \textbf{Reduces overfitting:} Increases effective data diversity without collecting new images
\end{itemize}

We empirically verified that prompt rotation improves transfer to novel prompt variations compared to training with a single fixed prompt.

\subsection{Inference Prompt Design}

For evaluation, we use a single, streamlined prompt that retains all critical instructions while using natural language. Table~\ref{tab:inference_prompt} presents our inference prompt.

\begin{table}[h]
\centering
\caption{\textbf{Inference prompt for evaluation.} This prompt is used consistently across all test evaluations, ablation studies, and model comparisons. It omits training-specific instructions (multi-clock selection, NO CLOCK fallback) while preserving core task requirements.}
\label{tab:inference_prompt}
\begin{tabular}{p{0.93\linewidth}}
\toprule
\textbf{Inference Prompt} \\
\midrule
\texttt{Find the most prominent analog clock in the image. The hour hand is the short, thick one, and the minute hand is the long, thin one. If a hand is between marks, choose the lower hour and the nearest minute. Your answer must be only in HH:MM format (e.g., 08:05).} \\
\bottomrule
\end{tabular}
\end{table}

\textbf{Simplification Rationale:}

\begin{itemize}[leftmargin=*,noitemsep]
    \item \textbf{Omitted instructions:} We remove "ignore digital displays," multi-clock selection details, and "NO CLOCK" fallback because our curated test set contains only valid analog clocks. This reduces prompt length and focuses the model's attention.
    
    \item \textbf{Explicit format example:} The phrase "e.g., 08:05" provides a concrete example reinforcing the expected output structure (leading zero, colon separator, no AM/PM).
    
    \item \textbf{Natural phrasing:} We use more conversational language ("The hour hand is...") compared to training prompts' terse notation ("hour = ..."). This tests whether the model has learned semantic understanding rather than pattern matching.
\end{itemize}

\subsection{Prompt Length Analysis}

Table~\ref{tab:prompt_length} compares token counts across different model tokenizers. Our prompts are designed to be concise yet comprehensive.

\begin{table}[h]
\centering
\caption{\textbf{Prompt length analysis across model tokenizers.} Token counts vary slightly due to different tokenization schemes (SentencePiece for Llama, tiktoken for Qwen, custom for Gemma). Inference prompt averages 68 tokens.}
\label{tab:prompt_length}
\begin{tabular}{lccc}
\toprule
\textbf{Prompt Type} & \textbf{Qwen} & \textbf{Llama} & \textbf{Gemma} \\
\midrule
Training Prompt A & 92 & 91 & 92 \\
Training Prompt B & 75 & 74 & 76 \\
Training Prompt C & 75 & 74 & 75 \\
\midrule
Inference Prompt & 68 & 66 & 69 \\
\bottomrule
\end{tabular}
\end{table}

The inference prompt's 66--69 token length strikes a balance between providing sufficient instruction and minimizing computational overhead. Shorter prompts ($<$30 tokens) lack critical guidance, while longer prompts ($>$150 tokens) provide diminishing returns while increasing latency.

\subsection{Cross-Model Consistency}

We use \textit{identical prompts} across all three VLM backbones (Qwen2.5-VL-7B, Llama-3.2-11B, Gemma3-12B) to ensure fair comparison. This design choice isolates the effect of model architecture and training procedure from prompt engineering, ensuring that performance differences reflect genuine model capabilities rather than prompt tuning artifacts. Any model-specific prompt optimization would confound our analysis and reduce reproducibility.

\subsection{Impact on DPO Training}

During DPO training, we use the same inference prompt for both chosen ($y_w$) and rejected ($y_l$) responses. This ensures that preference learning focuses on output quality rather than prompt interpretation differences. The consistent prompt also allows the Swap-DPO mechanism to function correctly—both the correct time and the geometrically swapped time are generated in response to identical instructions, isolating hand role confusion as the sole difference.



{
    \small
    \bibliographystyle{ieeenat_fullname}
    \bibliography{ref}
}


\end{document}